\documentclass[lettersize,journal]{IEEEtran}
\usepackage{amsmath,amsfonts}
\usepackage{algorithmic}
\usepackage{algorithm}
\usepackage{array}
\usepackage[caption=false,font=normalsize,labelfont=sf,textfont=sf]{subfig}
\usepackage{textcomp}
\usepackage{stfloats}
\usepackage{url}
\usepackage{verbatim}
\usepackage{graphicx}
\usepackage{cite}
\usepackage{xcolor}
\usepackage{diagbox}
\usepackage{tabularx}
\allowdisplaybreaks[4]
\makeatletter
\renewcommand{\maketag@@@}[1]{\hbox{\m@th\normalsize\normalfont#1}}%
\makeatother
\begin{document}

\title{Distributed Robust Learning based Formation Control of Mobile Robots based on Bioinspired Neural Dynamics}


\markboth{IEEE Transactions on Intelligent Vehicles}%
{Shell \MakeLowercase{\textit{et al.}}: A Sample Article Using IEEEtran.cls for IEEE Journals}

\author{Zhe Xu,~\IEEEmembership{Member,~IEEE,} Tao Yan,~\IEEEmembership{Graduate Student Member,~IEEE,} \\Simon X. Yang,~\IEEEmembership{Senior Member,~IEEE,} S. Andrew Gadsden,~\IEEEmembership{Senior Member,~IEEE} \\Mohammad Biglarbegian,~\IEEEmembership{Senior Member,~IEEE}
\thanks{This paper is accepted by \textit{IEEE Transactions on Intelligent Vehicles}, DOI: 10.1109/TIV.2024.3380000\\This work was supported by the Natural Sciences and Engineering Research Council (NSERC) of Canada. (\textit{Corresponding author: Simon X. Yang}) } 
\thanks{Zhe Xu and S. Andrew Gadsden are with the Intelligent and Cognitive Engineering (ICE) Lab, McMaster University, Hamilton, Ontario, Canada. e-mail: \{xu804; gadsden\}@mcmaster.ca}
\thanks{Tao Yan, and Simon X. Yang are with the Advanced Robotics and Intelligent Systems
(ARIS) Laboratory, School of Engineering, University of Guelph, Guelph, Ontario, Canada. e-mails: \{tyan03; syang\}@uoguelph.ca}
\thanks{Mohammad Biglarbegian is with the Department of Mechanical and Aerospace Engineering, Carleton University, Ottawa, Ontario, Canada. e-mail: mohammadbiglarbegian@cunet.carleton.ca}
}
\maketitle

\begin{abstract}
This paper addresses the challenges of distributed formation control in multiple mobile robots, introducing a novel approach that enhances real-world practicability. We first introduce a distributed estimator using a variable structure and cascaded design technique, eliminating the need for derivative information to improve the real time performance. Then, a kinematic tracking control method is developed utilizing a bioinspired neural dynamic-based approach aimed at providing smooth control inputs and effectively resolving the speed jump issue. Furthermore, to address the challenges for robots operating with completely unknown dynamics and disturbances, a learning-based robust dynamic controller is developed. This controller provides real time parameter estimates while maintaining its robustness against disturbances. The overall stability of the proposed method is proved with rigorous mathematical analysis. At last, multiple comprehensive simulation studies have shown the advantages and effectiveness of the proposed method.
\end{abstract}

\begin{IEEEkeywords}
Distributed estimation, Formation control, Mobile robot, Bioinspired neural dynamics, Learning process.
\end{IEEEkeywords}

\section{Introduction}
\IEEEPARstart{W}{ith} the advancement in vehicular technologies, mobile robots have found extensive applications across diverse fields \cite{Cao2023SafeRobots,Kim2023Energy-TimeRobots,Sun2016TwoRobots}. Among the various research topics in mobile robotics, cooperation has attracted significant attention. Within this area, the formation control is one of the fundamental research topics. The basic goal of formation control is to design formation protocols that allow robots to reach certain velocities or accelerations that maintain a prescribed relative position.

There have been many studies related to this formation protocol with different approaches such as virtual structure \cite{Zhou2018AgileStructures,Yang2023ResearchStructure}, behavior based \cite{Wen2023Behavior-BasedComputing,Feng2023DistributedCommunication}, artificial potential field \cite{Hu2023PotentialAvoidance,Souza2022ModifiedEnvironments}, and leader-follower based approach \cite{Xu2023DistributedFilter,Dai2020AdaptivePerformance,Lu2020Neuro-AdaptiveRobots}. Among all these formation approaches, it is noticed that the leader-follower approach has been the dominant method in real-world applications due to its relatively easy implementation, real-time performance, and good flexibility. 
However, conventional leader-follower approaches involve extensive centralized computations, becoming impractical as the number of robots increases\cite{Peng2013Leader-followerApproach,Miao2022Low-ComplexityFeedback}. Thus, the distributed approach has become a more promising approach in recent years \cite{Yan2022ConsensusControl,Zhang2023ParetoRobots,Shao2022DistributedSampling,Yao2023GuidingRobots}. In \cite{Liu2023DistributedApproach}, a distributed formation control of mobile robots from Udwadia-Kalaba approach was proposed. In \cite{Lu2020FormationEstimators}, distributed joint disturbance-and-leader estimators are designed to solve perturbed system dynamics. Similarly, \cite{Moorthy2022DistributedApproach} proposed a distributed estimation technique that was developed to estimate the leader's information.  By taking advantage of distributed techniques, it is seen that all these methods do not need to directly obtain the leader's state, thus offering more scalability and robustness.

On the other hand, the formation tracking control problem is vital for multiple nonholonomic mobile robots and extensive research has been studied on this subject \cite{Liu2023FormationTechnique,Lin2021AdaptiveConstraints,Liu2023SecureAttacks}. In \cite{Dai2020AdaptivePerformance}, they proposed an adaptive formation control of robots with prescribed transient and steady-state performance. Another work proposed a novel forward motion controller to realize tracking control of nonholonomic robots \cite{Peng2020MobileVehicles}. In Liu et al.'s work, a formation control of mobile robots based on an embedded control technique was developed to simplify the designing difficulties\cite{Liu2023FormationTechnique}, however, they only consider the kinematics while the dynamics is important to ensure the control efficiency as well.  Li, et, al. proposed a tube-based model predictive control method that is capable of providing a relatively smooth control input without speed jump issue \cite{Li2020RobustFormation}; however, the fundamental issue of model predictive control is that this method requires a large amount of computational power, which is a crucial factor to ensure real time performance. All of these works have made good contributions to the formation control of robots. However, robots may not start at their desired postures, these issues of posture discrepancies among robots at the initial stage have not been well studied. The consideration of this initial tracking error is critical because controllers could send large torque demand which may not be possible due to the power limitation that the motor can generate. In the better-case scenario, the controller may be saturated, which is undesirable in real world applications. Furthermore, the velocity constraint is another critical issue to be considered that ensures operational safety, control stability, and easy mechanical wearing. The control approaches that are capable of addressing this issue, such as model predictive control, often require a large number of calculations, making these methods hard to ensure real-time performance. 

Furthermore, the real dynamics of mobile robots are difficult to obtain, and there are two primary approaches to address such an issue. One is the robust control approach, another one is the learning based approach. In \cite{Wang2021DistributedAttacks}, a distributed adaptive control method is proposed that considers deception attacks. However, their work considers the dynamic of robots are given accurately, while the unconstructed dynamics is a critical factor in control design. Dai, et al. \cite{Dai2020AdaptivePerformance} considered the parameter uncertainties, however, their approach lacks the analysis under disturbances while the disturbances are unavoidable in real world applications. Another work that is developed in \cite{Lashkari2020DevelopmentCapability} considered the robustness of the control method with proved stability under disturbances and uncertainties. However, if their estimation of system parameters largely deviates from real value, the performance of their method will be limited. It is found that robust control methods often come at the price of losing the formation performance, and learning based methods lack the analysis of the robustness. 

Inspired by the previous works, it is noticed that there is a pressing need to develop a distributed formation control protocol for mobile robots operating under disturbances with completely unknown dynamics. In addition, Multiple challenges, including the speed jump, velocity constraints, and control robustness under disturbances should also be carefully addressed in the formation protocol design. Therefore, this paper aims to develop a novel distributed formation control method for mobile robots to address these aforementioned issues. The main contributions of this study are listed as follows 

1. A formation control method is developed based on a distributed estimation using cascade and variable structure design techniques. Compared with \cite{Moorthy2022DistributedApproach,Miao2018DistributedRobots,Xu2023DistributedFilter}, the derivative information is avoided in the positional estimation using a cascade design structure to reduce the computational complexity that improves the real time performance. Then, the velocity estimation is developed using a variable structure method. 

2. To address the speed jump issue and velocity constraints, a kinematic control method is developed to tackle the aforementioned issues by utilizing the special input-output properties of the bioinspired neural dynamics. Then, to allow better performance for the mobile robots operating under disturbances and unknown dynamics, a robust learning based dynamic control method is developed that actively provides real time parameter estimation of mobile robot dynamics and robustness against disturbances.

3. The stability of the overall formation protocol is rigorously provided using the Lyapunov stability theory. Numerous studies have demonstrated the proposed distributed bioinspired control method shows apparent advantages for mobile robots subject to velocity constraints and unknown dynamics.

The rest of the paper is organized as follows: Preliminaries and problem statements are demonstrated in Section II. Following this, Section III designs the overall distributed formation control method that includes distributed estimation, bioinspired kinematic control, learning based robust dynamic control. Then, multiple simulation studies are given in Section IV to demonstrate the superiority of the proposed method. Finally, Section V gives a conclusion of this work and leads to some potential directions for future works.

\section{Preliminaries and problem statement}
\subsection{Algebraic Graph Theory}
To model the communication topology of multiple mobile robots, the graph theory is adopted. Consider a network that consists of $n$ followers; the digraph is defined as $\mathcal{G} = \{ V,E\} $ with a node set $V=\{1,2,...,n\}$ and an edge set $E \subseteq V \times V$. A directed edge $(i,j) \in E$ refers to the access from node $i$ to node $j$. Then, the adjacency matrix is defined $A ={[{a_{ij}}]_{n \times n}}$. If $A$ is a symmetric matrix, then the graph is undirected. The elements in $A$ are defined as follows ${a_{ij}}=1$ if $i,j\in E$ and  ${a_{ij}}=0$ otherwise. Furthermore, $a_{ii}$ is assumed to be $0$. The Laplacian matrix $L = {[{l_{ij}}]_{n \times n}}$, which is incorporated with $A$ is defined as\par
\vspace{-2ex}\begin{small}\begin{equation}
    {l_{ij}} = \left\{ {\begin{array}{*{20}{c}}
{ - {a_{ij}}}\qquad \qquad i \ne j\\
{\sum\limits_{i = 1,i \ne j}^n {{a_{ij}}} }\qquad i = j.
\end{array}} \right.
\end{equation}\end{small}\normalsize

In this study, the leader is subscript by $r$ while the followers are given labels from $1$ to $n$. Denote $a = \left[ 
{{a_{1r}}},{{a_{2r}}},{...},{{a_{nr}}}
 \right]^T$ and $a_{ir}=1$ if tht leader is connected to the follower $i$, else $a_{ir}=0$. Then, we define $H \in {R^{n \times n}}$ as
\begin{equation}
    H = L + {\rm{diag}}(a). \label{2}
\end{equation}
Based on the previous studies, this paper introduces the following assumption and lemmas \cite{Moorthy2022DistributedApproach,Miao2018DistributedRobots, Li2020ReinforcementRobots}.

\noindent{\bf Assumption 1.}
The undirected graph $\mathcal{G}$ is connected, and there is at least one follower has access to the leader's state information, i.e., $a_{ir} = 1$.

\noindent{\bf Lemma 1.}
The matrix $H$ is positive definite if the undirected graph $\mathcal{G}$ is connected and at least one follower is the neighbor of the leader.

\noindent{\bf Lemma 2} \cite{Moorthy2022DistributedApproach}{\bf.} Let $V$: $\mathbb{R}^+\to\mathbb{R}^+$ be continuously differentiable and $W$: $\mathbb{R}^+\to\mathbb{R}^+$ be uniformly continuous and satisfy that for $\forall t \ge 0$ 
\begin{equation}
    \dot V(t) \le  - W(t) + {p_1}(t)V(t) + {p_2}(t)\sqrt {V(t)},
\end{equation}
where both ${p_1}$ and ${p_1}$ are nonnegative and belong to the $L_1$ space. Then, $V(t)$ is bounded, and there exists a constant $c$ such that $W(t)\to0$ and $V(t)\to c$ as $t\to \infty$.

\subsection{Problem Formulation}
The kinematic equations governing the motion of the leader and $i$th follower robot are as follows:
\begin{equation}
\begin{aligned}
    &{{\dot x}_{r}} = {\upsilon _{r}}\cos {\theta _{r}}, \quad
{{\dot y}_{r}} = {\upsilon _{r}}\sin {\theta _{r}}, \quad
{{\dot \theta }_{r}} = {\omega _{r}},\\
    &{{\dot x}_{i}} = {\upsilon _{i}}\cos {\theta _{i}}, \quad
{{\dot y}_{i}} = {\upsilon _{i}}\sin {\theta _{i}}, \quad
{{\dot \theta }_{i}} = {\omega _{i}},
\end{aligned}
\end{equation}
where subscript $i$ represents the follower $i$; ${P_r} = {\left[
{{x_r}},{{y_r}},{{\theta _r}} \right]^T}$ and ${P_i} = {\left[ 
{{x_i}}, {{y_i}}, {{\theta _i}} \right]^T}$ are respectively the leader and $i$th follower's positional state in $x$, $y$, and orientation in global frame; the linear velocities of the leader and follower $i$ in body fixed frame are respectively represented by ${\upsilon _{r}}$, ${\upsilon _{i}}$, ${\omega _{r}}$ and ${\omega _{i}}$. Furthermore, we introduce the following assumption for the leader robot.

\noindent{\bf Assumption 2} \cite{Miao2018DistributedRobots}{\bf.}
Denote the virtual leader's linear and angular velocities as $\upsilon_{r}$ and $\omega_{r}$, and assume they are bounded. Additionally, assume that $\dot \upsilon_r$ and $\dot \omega_r$ are also bounded and there exist constants ${\gamma _1}$ and ${\gamma _2}$ such that 
\begin{equation}
   \left[ {\begin{array}{*{20}{c}}
{\max \left| {{{\dot \upsilon }_r}(t)} \right|}\\
{\max \left| {{{\dot \omega }_r}(t)} \right|}
\end{array}} \right] = \left[ {\begin{array}{*{20}{c}}
{{\gamma _1}}\\
{{\gamma _2}}
\end{array}} \right].
\end{equation}

It is considered that each mobile robot shares a similar structure, and its dynamics is provided as
\begin{equation}
    \bar M(P_{i}){{\dot z }_{i}} = \bar B(P_{i})\tau_{i}  - \bar C(P_{i},\dot P_{i}){z _{i}}+{d_i},
\end{equation}
where ${{\xi }_{i}}=[{\upsilon _{i}}, {\omega _{i}}]^T$; $\bar M(P_{i})$ is the inertial matrix; $\bar B(P_{i})$ is the transformation matrix; $\bar C(P_{i},\dot P_{i})$ is the centrifugal and Coriolis matrix of the $i$-th mobile robot; and the disturbances are denoted by ${{d_i}} = {\left[ 
{{d _{1i}}}, {{d _{2i}}}
 \right]^T}$ and these disturbances satisfy the following assumption

\noindent{\bf Assumption 3} \cite{Li2020ReinforcementRobots}{\bf.}
It is assumed that the disturbances acting on the mobile robot are bounded such that 
\begin{equation}
\left[ {\begin{array}{*{20}{c}}
{\max \left| {{d_{1i}}(t)} \right|}\\
{\max \left| {{d_{2i}}(t)} \right|}
\end{array}} \right] = \left[ {\begin{array}{*{20}{c}}
{{\psi _{1i}}}\\
{{\psi _{2i}}}
\end{array}} \right],
\end{equation}
where $\psi _{1i}$ and $\psi _{2i}$ are positive constants. 

In the leader-follower formation control of mobile robots, each follower robot is programmed to maintain a specific relative position with respect to the leader. This relative positioning for the $i$-th follower is defined by $\left[ {\Delta {x_{i}}}, {\Delta {y_{i}}} \right]$, where ${\Delta {x_{i}}}$ and ${\Delta {y_{i}}}$ are the relative position to the leader in the inertial frame. Furthermore,  followers are aligned in a manner that mirrors the orientation of the leader, ensuring a coherent and structured group movement. 

Therefore, the objective of this manuscript is to develop a distributed formation control strategy that ensures the formation of multiple mobile robots is maintained such that
\begin{equation}
    \mathop {\lim }\limits_{t \to \infty } \left( {{x_r} - {x_i}} \right) = \Delta {x_i},
\end{equation}
\begin{equation}
   \mathop {\lim }\limits_{t \to \infty } \left( {y_{r}} - {y_{i}}\right)  =  \Delta {y_{i}},
\end{equation}
\begin{equation}
    \mathop {\lim }\limits_{t \to \infty } \left({\theta_{r}} - {\theta_{i}}\right) = 0,
\end{equation}
where $\tau _i^x$, $\tau _i^y $, and $\tau _i^\theta $ are arbitrary small thresholds.
\vspace{-8pt}
\section{Formation Protocol Design}
This section designs the formation protocol that enables the group of mobile robots to achieve its prescribed formation as well as track its desired trajectory. First, a distributed estimation process is introduced that allows each individual robot to estimate the leader's positional and velocity state information. Following the distributed estimator design, a bioinspired neural dynamic based kinematic control is subsequently proposed that addresses the velocity constraints while providing smooth velocity control input. Then, a learning based robust dynamic controller is developed with unknown system parametric information.

\subsection{Distributed Estimator Design}
First, the distributed estimator is developed using the consensus estimation error. Each individual estimates the leader's posture and velocities using the available neighboring information. This design utilizes the cascade design approach, and the following processes demonstrate the development of the approach.
 
 Define the consensus estimation error of $i$th mobile robot ${\tilde P_{i}} = {\left[ 
{{\tilde x}_{ir}}, {{\tilde y}_{ir}}, {{\tilde \theta }_{ir}}
 \right]^T}$ as 
 \begin{equation}
     {\tilde P_i} = \sum\limits_{j = 1}^n {{a_{ij}}\left( {{ P_{ir}} - { P_{jr}}} \right) + {a_{ir}}{e_{ir}}},   \label{11}
 \end{equation}
 where ${{P}_{ir}} = {\left[ {{x_{ir}}},{{ y_{ir}}},{{ \theta _{ir}}} \right]^T}$ and ${{ P}_{jr}} = {\left[ {{ x_{jr}}},{{ y_{jr}}},{{ \theta _{jr}}} \right]^T}$ are the estimation of leader's posture of robot $i$ and $j$, respectively; and ${e_{ir}} = { {P}_{ir}} - {P_r}$. Based on the defined estimation error, the following estimation process is designed for $i$th mobile robot to estimate the leader's posture as 
 \begin{equation}
     {\dot { {P}}_{ir}} = \left[ {\begin{array}{*{20}{c}}
{{\dot { x}_{ir}}}\\
{{\dot { y}_{ir}}}\\
{{\dot { \theta} _{ir}}}
\end{array}} \right] = \left[ {\begin{array}{*{20}{c}}
{{{\upsilon} _{ir}}\cos {\theta _{ir}} - {k_{xi}}{\tilde x}_{ir}}\\
{{{\upsilon} _{ir}}\sin {\theta _{ir}} - {k_{yi}}{\tilde y}_{ir}}\\
{{{\omega} _{ir}} - {k_{\theta i}}{\tilde \theta }_{ir}}
\end{array}} \right],\label{12}
 \end{equation}
 where ${k_{pi}} = {\left[ {{k_{xi}}},{{k_{yi}}},{{k_{\theta i}}}\right]^T}$ is a positive design vector; variables ${ \upsilon _{ir}}$ and ${ \omega _{ir}}$ are respectively the estimated leader's linear and angular velocities of $i$th follower. From here, we define the neighborhood velocity estimation errors as 
 \begin{equation}
\begin{array}{l}
{e_{i\upsilon }} = \sum\limits_{j = 1}^n {{a_{ij}}\left( {{ \upsilon _{ir}} - { \upsilon _{jr}}} \right) + {a_{ir}}{e_{i\alpha }}}, \\
{e_{i\omega }} = \sum\limits_{j = 1}^n {{a_{ij}}\left( {{ \omega _{ir}} - { \omega _{jr}}} \right) + {a_{ir}}{e_{i\beta }}} ,
\end{array}
 \end{equation}
where ${e_{i\alpha }} = { \upsilon _{ir}} - {\upsilon _r}$ and ${e_{i \beta }} = {\omega _{ir}} - {\omega _r}$. We design the velocity estimator using a variable structure method and propose the following estimation process as
\begin{equation}
    \begin{array}{l}
{{\dot { \upsilon} }_{ir}} =  - {k_{a1i}}{\rm{sgn}}\left( {{e_{i\upsilon }}} \right) - {k_{b1i}}{e_{i\upsilon }},\\
{{\dot { \omega} }_{ir}} =  - {k_{a2i}}{\rm{sgn}}\left( {{e_{i\omega }}} \right) - {k_{b2i}}{e_{i\omega }},
\end{array}\label{13}
\end{equation}
where $\rm{sgn}\left(.\right)$ represents the sign function; $k_{a1i}$, $k_{a2i}$, $k_{b1i}$ and $k_{b2i}$ are all positive design constants.
Thus, the distributed estimator design protocol is completed. 

\noindent{\bf Remark 1} Compared to some of the previous studies \cite{Moorthy2022DistributedApproach,Miao2018DistributedRobots}, the proposed posture estimators do not require derivative information from the neighbors. Furthermore, the velocity estimator uses a variable structure approach to design the distributed estimator yet also without needing the velocity derivative information. Then, we introduce the following theorem to prove the stability and convergence of the proposed distributed estimators.           

\noindent{\bf Theorem 1} The proposed distributed estimators given in \eqref{12} and \eqref{13} are asymptotically stable if Assumption 1 holds, such that ${e_{ir}} \to 0$ and ${e_{i\upsilon}}\to 0$ and ${e_{i\omega}}\to 0$ as time$\to \infty$.

\noindent{{\textit{Proof}}}
The proposed posture estimator uses information that is obtained by the velocity estimator, therefore, we first need to prove the convergence of the velocity estimation error. 

First, define ${e_\alpha } = {\left[ 
{{e_{1\alpha }}},{...},{{e_{n\alpha }}}
 \right]^T}$ and ${e_\upsilon } = {\left[ 
{{e_{1\upsilon }}},{...},{{e_{n\upsilon }}} \right]^T}$. Then, it is relatively easy to verify that
\begin{equation}
    {e_\upsilon } = H{e_\alpha }.\label{14}
\end{equation}
Here, we define ${k_{a1}} = {\rm{diag}}\left( 
{{k_{a11}}}, {...}, {{k_{a1n}}}
 \right)$, ${k_{b1}} = {\rm{diag}}\left( {{k_{b11}}}, {...}, {{k_{b1n}}} \right)$, and ${\hat \upsilon _r} = \left[ {{{ \upsilon }_{1r}}}, {...},{{{\upsilon }_{nr}}}  \right]^T$. Then, we propose the Lyapunov candidate function as ${V_{1}} = \frac{1}{2}e_v^TH^{-1}{e_v}$. Combining with \eqref{14}, the time derivative of $V_{1}$ is calculated as
\begin{equation}
\begin{aligned}
   {{\dot V}_1}{\text{ }} &= e_\upsilon ^T{H^{ - 1}}{{\dot e}_\upsilon } = e_\upsilon ^T{H^{ - 1}}H{{\dot e}_\alpha } = e_\upsilon ^T\left( {{{\dot {\hat {\upsilon }}_r}} - {{\dot \upsilon }_r}{1_n}} \right)\\
 &=  - e_\upsilon ^T{k_{a1}}{e_\upsilon } - e_\upsilon ^T{k_{b1}}{\rm{sgn(}}{e_\upsilon }) - e_\upsilon ^T{{\dot \upsilon }_r}{1_n}, \label{15}
\end{aligned}
\end{equation}
where ${\rm{sgn(}}{e_\upsilon }) = {\left[ 
{{\rm{sgn(}}{e_{1\upsilon }})},{...},{{\rm{sgn(}}{e_{n\upsilon }})}
 \right]^T}$. We recall Assumption 2, $\dot V_1$ has the following inequalities
\begin{equation}
{{\dot V}_1} \le  - {\lambda _1}\left\| {{e_\upsilon }} \right\|^2 - {\lambda _2}{\left\| {{e_\upsilon }} \right\|_1} + {\gamma _1}{\left\| {{e_\upsilon }} \right\|_1},\label{16}
\end{equation}
where $\lambda_1$ and $\lambda_2$ are respectively the minimum eigenvalue of $k_{a1}$ and $k_{b1}$. Therefore, by ensuring $\lambda_2\ge\gamma_1$, we have ${{\dot V}_1} \le  - {\lambda _1}\left\| {{e_\upsilon }} \right\|^2$. Here, we have proved the asymptotic convergence of the proposed control. Given that $H$ is a positive definite symmetric matrix, ${{e_\upsilon }} $ converges to zero exponentially.

However, if the parameters are not well tuned, the bound of the proposed method can still be ensured. First by applying Cauchy-Schwartz inequality, \eqref{16} is written as
\begin{equation}
\begin{aligned}
        {{\dot V}_1} \le&   - {\lambda _1}\left\| {{e_\upsilon }} \right\|^2 - {\lambda _2}{\left\| {{e_\upsilon }} \right\|} + {\gamma _1}\sqrt n {\left\| {{e_\upsilon }} \right\|}\\
         \le& - \left( {{\lambda _1} - {\iota _1}} \right)\left\| {{e_\upsilon }} \right\|^2, \quad {\rm{whenever}}\left\| {{e_\upsilon }} \right\| \ge \mu_1,\label{17}
\end{aligned}
\end{equation}
where $\iota_1\in(0,\lambda_1)$, and ${\mu _1} = {{\left( {{\gamma _1}\sqrt n  - {\lambda _2}} \right){\rm{ }}} \mathord{\left/
 {\vphantom {{\left( {{\gamma _1}\sqrt n  - {\lambda _2}} \right){\rm{ }}} {{\iota _1}}}} \right.
 \kern-\nulldelimiterspace} {{\iota _1}}}$.
 
 Using a similar approach as shown from \eqref{15} to \eqref{17}, we proposed the Lyapunov candidate function for the angular velocity estimator as ${V_{2}} = \frac{1}{2}e_\omega^TH^{-1}{e_\omega}$. Then, after some calculations, the time derivative of $V_2$ is calculated as
 \begin{equation}
 \begin{aligned}
 \dot V_2&= - {\lambda _3}\left\| {{e_\omega }} \right\|^2 - {\lambda _4}{\left\| {{e_\omega }} \right\|_1} + {\gamma _2}{\left\| {{e_\omega }} \right\|_1}\\
   & \le  - {\lambda _3}\left\| {{e_\omega }} \right\|^2 - {\lambda _4}{\left\| {{e_\omega }} \right\|} + {\gamma _2}\sqrt n {\left\| {{e_\omega }} \right\|}\\
   &\le - \left( {{\lambda _3} - {\iota _2}} \right){\left\| {{e_\omega }} \right\|^2},\quad {\rm{whenever}}\left\| {{e_\omega }} \right\| \ge {\mu _2},
 \end{aligned}\label{19}
 \end{equation}
 where $\iota_2\in(0,\lambda_3)$; $\lambda_3$ and $\lambda_4$ are respectively the minimum eigenvalue of $k_{a2}$ and $k_{b2}$; and ${{{\mu _2} = \left( {{\gamma _2}\sqrt n  - {\lambda _4}} \right)} \mathord{\left/
 {\vphantom {{{\mu _2} = \left( {{\gamma _2}\sqrt n  - {\lambda _4}} \right)} {{\iota _2}}}} \right.
 \kern-\nulldelimiterspace} {{\iota _2}}}$. Once again, based on the first line of \eqref{19}, if we can ensure $\lambda_4$ is greater than ${{\gamma _2}\sqrt n }$, $e_\omega$ converges to zeros exponentially, in the worst case scenario, whenever ${\left\| {{e_\omega }} \right\|} \ge {\mu _2}$, the input to state stability can be ensured.

 In order to prove the stability of the proposed positional estimator, we first define $\tilde P = {\left[ 
{\tilde P_1^T},{...},{\tilde P_n^T}
 \right]^T}$, ${e_r} = {\left[ 
{{e_{1r}}^T},{...},{{e_{nr}}^T} \right]^T}$, ${k_p} = {\rm{diag}}\left( 
{{k_{p1}}},{...},{{k_{pn}}}\right)$, and $\tilde \upsilon_r  = {\left[ 
{\tilde \upsilon _{r1}^T},{...},{\tilde \upsilon _{rn}^T}
 \right]^T}$ with $\tilde \upsilon_{ri}$ is defined as
\begin{equation}
        {{\tilde \upsilon }_{ri}} = \left[ {\begin{array}{*{20}{c}}
{{\upsilon _{ir}}\cos {\theta _{ir}} - {{\dot x}_{r}}}\\
{{\upsilon _{ir}}\sin {\theta _{ir}} - {{\dot y}_{r}}}\\
{{\omega _{ir}} - {{\dot \theta }_{r}}}
\end{array}} \right].\label{20}
\end{equation}
To ensure that $\tilde \upsilon_i$ is bounded, denote ${\tilde \theta _r} = {\left[ 
{{{\tilde \theta }_{1r}}},{...},{{{\tilde \theta }_{nr}}} \right]^T}$ and ${e_\beta } = {\left[ 
{{e_{1\beta }}},{...},{{e_{n\beta }}}
 \right]^T}$, we first propose the Lyapunov candidate function as ${V_3} = \frac{1}{2}\tilde \theta _r^TH^{-1}{{\tilde \theta }_r}$ and combine with \eqref{12}, its time derivative is calculated as 
\begin{equation}
    \begin{aligned}
       {{\dot V}_3} =&  - \tilde \theta _r^T{k_\theta }{{\tilde \theta }_r} + \tilde \theta _r^T{e_\beta }\\
     \le &  - \left( {{\lambda _5} - {\iota _3}} \right){\left\| {{{\tilde \theta }_r}} \right\|^2},\quad {\rm{whenever}}\left\| {{{\tilde \theta }_r}} \right\| \ge  {{\mu_3}}, 
    \end{aligned}
\end{equation}
where $\iota_3\in(0,\lambda_5)$; $\lambda_5$ is the minimum eigenvalue of ${k_\theta } = {\rm{diag}}\left( 
{{k_{1\theta }}}, {...}, {{k_{n\theta }}}
 \right)$; and ${\mu _3} = {{\left\| {{e_\beta }} \right\|} \mathord{\left/
 {\vphantom {{\left\| {{e_\beta }} \right\|} {{\iota _3}}}} \right.
 \kern-\nulldelimiterspace} {{\iota _3}}}$. Thus, we can imply that the error between $ \theta_{ir}$ and $\theta_{r}$ is bounded if ${\left\| {{e_\beta }} \right\|}$ is bounded, consequently, with the results obtained in \eqref{17} and \eqref{19}, we can conclude that $\tilde \upsilon_{ri}$ is bounded. Furthermore, based on the results obtained in \eqref{19}, $\left\| {H{e_\omega }} \right\|\to0$ exponentially, thus, \scalebox{0.7}{$\left\| {{{\tilde \theta }_r}} \right\|$} to zeros exponentially as well. Denote $\tilde P = {\left[ {\tilde P_1^T},{...},{\tilde P_n^T} \right]^T}$ and ${{\hat P}_r} = {\left[ 
{ P_{1r}^T},{...},{ P_{nr}^T}
 \right]^T}$. The corresponding Lyapunov candidate function that proves the positional estimator is proposed as ${V_4} = \frac{1}{2}{{\tilde P}^T}M^{-1}\tilde P$.
 \begin{equation}
     {{\dot V}_4} = {{\tilde P}^T}M^{-1}M{{\dot e}_r} = \tilde P^TM^{-1}M\left( {{{\dot {\hat P}}_r} - {{\dot P}_r}\otimes{1_n}} \right),\label{23}
 \end{equation}
where $M = {I_3} \otimes H$. By substituting the estimator designed in \eqref{12} into \eqref{23}, we obtain
\begin{equation}
\begin{aligned}
{{\dot V}_4} &=   {{\tilde P}^T}\tilde \upsilon_r  - {{\tilde P}^T}{k_p}\tilde P\\
             & \le  - \left( {{\lambda _6} - {\iota _4}} \right){\left\| {\tilde P} \right\|^2},\quad {\rm{whenever}}\left\| {\tilde P} \right\| \ge {\mu _4},\label{24}
\end{aligned}
\end{equation}
where $\iota_4\in(0,\lambda_6)$; $\lambda_6$ is the minimum eigenvalue of $k_p$; and ${\mu _4} = {{\left\| {\tilde \upsilon_r } \right\|} \mathord{\left/
 {\vphantom {{\left\| {\tilde \upsilon } \right\|} {{\iota _4}}}} \right.
 \kern-\nulldelimiterspace} {{\iota _4}}}$. It is obvious that $\tilde \upsilon_r\to0$ as well. Thus, the positional estimator is asymptotically stable at $\tilde P=0$.

\subsection{Bioinspired Backstepping Controller Design}
Followed by the distributed estimator design, a bioinspired neural dynamic based kinematic control approach is proposed using the estimated leader's state. The goal is to stabilize the tracking error between the leader's and the follower's posture. Then, this tracking error for $i$th mobile robot in the inertial frame is given as
\begin{equation}
    \begin{array}{l}
{e_{ix}} = { x_{ir}} - {x_{i}} - \Delta {x_i},\\
{e_{iy}} = { y_{ir}} - {y_{i}} - \Delta {y_i},\\
{e_{i\theta  }} = { \theta _{ir}} - {\theta _{i}},
\end{array}
\end{equation}
where ${e_{ix}}$, ${e_{iy}}$, and ${e_{i\theta}}$ are the tracking errors in $x$, $y$ directions, and orientation, respectively; $\Delta {x_i}$ and $\Delta {y_i}$ are the predefined relative position between follower $i$ and the leader in $x$ and $y$ directions, respectively. We have proved the convergence of estimated leader posture $P_{ir}$ and the actual leader $P_r$. Thus, the goal is to design a controller that drives the robots to maintain their relative position to the estimated leader position. Then, we can obtain the relations of the tracking error from the inertial frame to the body fixed frame as
\begin{equation}
    \left[ {\begin{array}{*{20}{c}}
{{{\tilde{x}_i}}}\\
{{{\tilde{y}_i}}}\\
{{{\tilde{\theta}_i }}}
\end{array}} \right] = \left[ {\begin{array}{*{20}{c}}
{\cos {\theta _{i}}}&{\sin {\theta _{i}}}&0\\
{ - \sin {\theta _{i}}}&{\cos {\theta _{i}}}&0\\
0&0&1
\end{array}} \right]\left[ {\begin{array}{*{20}{c}}
{{{e_{ix}}}}\\
{{{e_{iy}}}}\\
{{{e_{i\theta }}}}
\end{array}} \right],
\end{equation}
where ${{{\tilde{x}_i}}}$ and ${{{\tilde y_{i}}}}$ are driving and lateral tracking errors in the body fixed frame, respectively, and ${{{\tilde \theta_{ i }}}}$ is the tracking error of orientation. Through some of the calculations, we can obtain the error dynamics in the inertial frame as 
\begin{equation}
\left[ {\begin{array}{*{20}{c}}
{{{\dot {\tilde x}_{i}}}}\\
{{{\dot {\tilde y}_{i}}}}\\
{{{ \dot {\tilde\theta}_{i }}}}
\end{array}} \right] = \left[ {\begin{array}{*{20}{c}}
{{\omega _{i}}{{\tilde y}_{i}} - {\upsilon _{i}} + { \upsilon _{ir}}\cos {{\tilde \theta }_{i}} + {\Omega _{ix}}}\\
{ - {\omega _{i}}{{\tilde x}_{i}} + { \upsilon _{ir}}\sin {{\tilde \theta }_{i}}  + {\Omega _{iy}}}\\
{{ \omega _{ir}} - {\omega _{i}}+\Omega_{i\theta}}\label{29}
\end{array}} \right],
\end{equation}
with ${\Omega _{ix}}$, ${\Omega _{iy}}$, and ${\Omega _{i\theta}}$ are respectively defined as
\begin{equation}
    \begin{aligned}
{\Omega _{ix}} & = ({{\dot { x}}_{ir}} - {\upsilon _{ir}}\cos { \theta _{ir}})\cos {\theta _i} + ({{\dot{ y}}_{ir}} - {\upsilon _{ir}}\sin {\theta _{ir}})\sin {\theta _i},\\
{\Omega _{iy}} & = ({{\dot y}_{ir}} - {\upsilon _{ir}}\sin {\theta _{ir}})\cos {\theta _i} - ({{\dot x}_{ir}} - {\upsilon _{ir}}\cos {\theta _{ir}})\sin {\theta _i},\\
{\Omega _{i\theta}}& = \dot \theta_{ir}- \omega_{ir},\label{30}
    \end{aligned}
\end{equation}
Based on the error dynamics that is provided in \eqref{29}. We take advantage of the bioinspired neural dynamics to develop the controller using the backstepping design technique. The proposed kinematic control design is provided as
\begin{equation}
    \begin{aligned}
    {\upsilon^{cmd} _i} =& {\upsilon _{ir}}\cos {{\tilde \theta }_i} + {k_{1i}}{V_{si}},
    \\ \omega _i^{cmd} =& {\omega _{ir}} + {k_{2i}}{\upsilon _{ir}}{\tilde y_i} + {k_{3i}}{\upsilon _{ir}}\sin \tilde \theta_i, \label{31}
    \end{aligned}
\end{equation}
where ${\upsilon^{cmd} _i}$ and ${\omega^{cmd} _i}$ are respectively the linear and angular velocity command; ${k_{1i}}$, ${k_{2i}}$, and ${k_{3i}}$ are positive design constants; $V_{si}$ is the output from the bioinspired neural dynamics that is provided in \eqref{32}. This neural dynamic model is originally used to demonstrate the neural activity of a membrane model \cite{Hodgkin1952ANerve}.

\noindent{\bf Remark 2.} In conventional control design, if initial tracking errors are not zeros, term ${k_{1i}}{{\tilde x}_i}$ will cause speed jump issues that make the demanding torque tend to be infinitely large. This issue is avoided with the implementation of bioinspired neural dynamics. 

This neural dynamics is defined as
\begin{equation}
    {\dot V_{si}} =  - {A_i}{V_{si}} + \left( {{B_i} - {V_{si}}} \right)f\left( {{{\tilde x}_{i}}} \right) - \left( {{D_i} + {V_{s1}}} \right)g\left( {{{\tilde x}_{i}}} \right),\label{32}
\end{equation}
where $A_i$, $B_i$, and $D_i$ are the positive design constants that represent the passive decay rate, and the excitatory and inhibitory inputs, respectively. By setting $B_i=D_i$, we can further rewrite \eqref{32} into the following form as
\begin{equation}
    \begin{aligned}
{{\dot V}_{si}} =  - {\Lambda _{i}}{V_{si}} + {B_{i}}{{U}_{i}},
    \end{aligned}\label{33}
\end{equation}
where ${\Lambda _{i}} = {A_{i}} + \left| {{U_{i}}} \right|$ with ${U_{i}}={\tilde x}_{i}$. 

\noindent{\bf Remark 3.} This bioinspired neural dynamics model that is provided in \eqref{33} has several advantages. First, this model has a bounded output regardless of the input, such that the output is always bounded between $(-D_{i}, B_{i})$. Taking this characteristic of this model, we have the maximum linear velocity control command is ensured to be bounded by $\upsilon_{ir}+k_{1i}B_{i}$, thus addressing the velocity constraint issue.

\noindent{\bf Remark 4.}
This dynamics model acts more like a first-order low pass filter regarding ${{{B_{i}}{U_{i}}} \mathord{\left/
 {\vphantom {{{B_{i}}{U_{i}}} {{\Lambda _{i}}}}} \right.
 \kern-\nulldelimiterspace} {{\Lambda _{i}}}}$ as the input and $1/\Lambda_{i}$ as its time constant. As such, this filtering capability offers extra smoothness to the control command when considering noises. Secondly, the dynamic process of the bioinspired model is dynamic, which makes the demanding initial torque start from zero. 

\subsection{Learning based robust controller design}
This subsection designed a dynamic learning based robust dynamic sliding mode controller. In real world applications, the dynamics of the mobile robot are difficult to obtain. However, obtaining an accurate dynamic model, which poses substantial challenges when designing controllers, is critical to ensure control efficiency. Therefore, this part designs an online learning procedure that actively estimates the system parameters with limited state information. Using the estimated parameter information, a robust controller is then proposed.

First, we may rewrite the robot dynamics into the following form as
\begin{equation}
  {\dot z_i} = \left[ {\begin{array}{*{20}{c}}
{{\dot\upsilon _i}}\\
{{\dot\omega _i}}
\end{array}} \right] = \left[ {\begin{array}{*{20}{c}}
{{\tau _{ai}}}&0\\
0&{{\tau _{bi}}}
\end{array}} \right]\left[ {\begin{array}{*{20}{c}}
{{a_i}}\\
{{b_i}}
\end{array}} \right] + \left[ {\begin{array}{*{20}{c}}
{{d_{1i}}}\\
{{d_{2i}}}
\end{array}} \right],\label{dy}
\end{equation}
where ${\tau _{ai}} = {\tau _{Li}} + {\tau _{Ri}}$ and ${\tau _{bi}} = {\tau _{Li}} - {\tau _{Ri}}$ with $\tau_{Li}$ and $\tau_{Ri}$ are the generated torque from left and right wheels, respectively; ${a_i} = {1 \mathord{\left/{\vphantom {1 {{m_i}{r_i}}}} \right.\kern-\nulldelimiterspace} {{m_i}{r_i}}}$ and ${b_i} = {{{l_i}} \mathord{\left/{\vphantom {{{l_i}} {{I_i}{r_i}}}} \right.\kern-\nulldelimiterspace} {{I_i}{r_i}}}$, where $m_i$ is the mass of the robot; $I_i$ is the moment of inertia, $r_i$ is the radius of the driving wheel, $l_i$ is the azimuth length from centre of gravity to the driving wheel. It is assumed that both $a_i$ and $b_i$ are unknown, therefore, we need to design the estimation protocol that actively provides the real time estimation of these parameters. We define the estimation error as 
\begin{equation}
    {S_i} = {k_{4i}}\int {{e_{vi}}}  + {e_{vi}},\label{42}
\end{equation}
where $k_{4i}$ is a positive definite design matrix and ${e_{vi}} = {\left[ 
{{{\hat \upsilon }_i} - {\upsilon _i}},{{{\hat \omega }_i} - {\omega _i}}
 \right]^T}$, with ${{\hat \upsilon }_i}$ and ${{\hat \omega }_i}$ are respectively the estimated linear and angular velocities. By taking the time derivative of \eqref{42}, we can calculate the following dynamics as
\begin{equation}
    {{\dot S}_i} = {k_{4i}}{e_{vi}} + {{\dot {\hat z}}_i} - {{\dot z}_i}.\label{43}
\end{equation}
Then, we design the following online process to estimate parameters $a_i$ and $b_i$.
\begin{equation}
    \begin{aligned}
        {{\dot {\hat z}}_i} &= {\tau _i}{{\hat c}_i} - {k_{4i}}{e_{vi}} - {k_{5i}}{S_i},\\
        {{\dot {\hat c}}_i} &= -{\tau _i}{k_{4i}}{S_i},\label{37}
    \end{aligned}
\end{equation}
where ${{\hat c}_i} = {\left[ 
{{{\hat a}_i}},{{{\hat b}_i}}
\right]^T}$ and ${{\hat a}_i}$ and ${{\hat b}_i}$ are respectively the estimation of $a_i$ and $b_i$. Note that this estimation process is online because it only requires the current measurement within the feedback loop, which barely depends on the historical data. Substituting the proposed estimator in \eqref{37} into \eqref{43}, we obtain the error dynamics of $\dot S_i$ as
\begin{equation}
\begin{aligned}
   {{\dot S}_i} &= {k_{4i}}{e_{\upsilon i}} + {\tau _i}{{\hat c}_i} - {k_{4i}}{e_{\upsilon i}} - {k_{5i}}{S_i} - {\tau _i}{c_i} - {d_i}\\
    &= {\tau _i}{{\tilde c}_i} - {k_{5i}}{S_i} - {d_i}.
\end{aligned}
\end{equation}
Then, we can further rewrite the dynamics of the mobile robot defined in \eqref{dy} as 
 \begin{equation}
     {{\dot z}_i} = \tau \hat c_i  - {k_{5i}}{S_i} - {{\dot S}_{i}}.
 \end{equation}
To further easy the analysis, we define $\Delta _i= {k_{5i}}{S_i} + {{\dot S}_{i}}$. Due to the effects of disturbances and unmodelled dynamics, the actual velocity is different from the velocity command that is generated by the kinematic controller. Then, we need to design a controller that ensures the convergence of $\tilde \upsilon_i$ and $\tilde \omega_i$, and the sliding mode based approach is a promising technique that is robust to these uncertainties. Then, using the estimated parameters along with obtained velocities, the robust learning based controller is designed as
\begin{equation}
{\tau _{L,i}} = \frac{1}{{2\hat a_i}}\left( {{{\dot \upsilon }^{cmd}}_i + {c_{ai}}{\mathop{\rm sgn}} ({{\tilde \upsilon }_i})} \right) - \frac{1}{{2\hat b_i}}({{\dot \omega }^{cmd}}_i + {c_{bi}}{\mathop{\rm sgn}} ({{\tilde \omega }_i})),\label{38}
\end{equation}
\begin{equation}
\begin{aligned}
{\tau _{R,i}} = \frac{1}{{2\hat a_i}}\left( {{{\dot \upsilon }^{cmd}}_i + {c_{ai}}{\mathop{\rm sgn}} ({{\tilde \upsilon }_i})} \right) + \frac{1}{{2\hat b_i}}({{\dot \omega }^{cmd}}_i + {c_{bi}}{\mathop{\rm sgn}} ({{\tilde \omega }_i})),
\end{aligned}\label{39}
\end{equation}
where ${c_{ai}}$ and ${c_{bi}}$ are the positive control parameters; ${{\tilde \upsilon }_i}=\upsilon^{cmd}_i-\upsilon_i$ and ${{\tilde \omega }_i}=\omega^{cmd}_i-\omega_i$ are respectively the error between the velocity command and actual velocity that is generated from the controller.

\subsection{Stability Analysis}
This subsection provides the overall stability of the closed loop system of the proposed method. We divided the stability analysis into two different parts.

\subsubsection{Step 1}First, to prove the stability of the dynamics controller. Define ${{\tilde c}_i} = {\left[ 
{{{\tilde a}_i}},{{{\tilde b}_i}} \right]^T}$ with ${{\tilde a}_i} = {{\hat a}_i} - {a_i}$ and ${{\tilde b}_i} = {{\hat b}_i} - {b_i}$, we present the following Lyapunov candidate function that firstly proves the convergence of the proposed learning process
\begin{equation}
   {V_{5i}} =\frac{1}{2}\sum\limits_{i=1}^n {\tilde c_i^T{{\tilde c}_i}}  + \frac{1}{2}\sum\limits_{i=1}^n {S_i^T{k_{4i}}{S_i}}.\label{46}
\end{equation}
The time derivative of \eqref{46} is calculated as
\begin{equation}
    \begin{aligned}
        {{\dot V}_{5i}} &=  - \sum\limits_{i=1}^n {\tilde c_i^T{\tau _i}{k_{4i}}{S_i}}  + \sum\limits_{i=1}^n {S_i^T{k_{4i}}\left( {{\tau _i}{{\tilde c}_i} - {k_{5i}}{S_i} - {d_i}} \right)} \\
         & = - \sum\limits_{i=1}^n {S_i^T{k_{4i}}{k_{5i}}{S_i}}  - \sum\limits_{i=1}^n {S_i^T{k_{4i}}{d_i}}  \\
         &\le -\sum\limits_{i=1}^n{{\lambda _{si}}{\left\| {{S_i}} \right\|^2}} + \sum\limits_{i=1}^n{\left\| {{k_{4i}}{d_i}} \right\|\left\| {{S_i}} \right\|}\\
         & \le  - \sum\limits_{i=1}^n {\left( {{\lambda _{si}} - {\iota _{5i}}} \right){{\left\| {{S_i}} \right\|}^2}}, \quad \text{whenever} \left\| {{S_i}} \right\| \ge {\mu _{5i}},\label{40}
    \end{aligned}
\end{equation}
where ${\lambda _{si}}$ is the minimum eigenvalue of $k_{4i}k_{5i}$; $\iota_{5i}$ is an arbitrary number between 0 and $\lambda_{si}$; and ${\mu _{5i}} = {{\left\| {{k_{4i}}{d_i}} \right\|} \mathord{\left/
 {\vphantom {{\left\| {{k_{4i}}{d_i}} \right\|} {{\iota _{4i}}}}} \right.
 \kern-\nulldelimiterspace} {{\iota _{5i}}}}$. Then, the stability of the proposed learning process under disturbances is input to state stable regarding the input $d_i$. Recalling Assumption 3, we have $d_i$ is bounded, Therefore, $\left\| {{S_i}} \right\|$ is bounded, and the estimation error is bounded as well. 
Then we propose the Lyapunov candidate function for the dynamic controller as
 \begin{equation}
     {V_{6}} =  \sum\limits_{i=1}^n{\frac{1}{2}\tilde z_i^T{{\tilde z}_i}},
 \end{equation}
where ${\tilde z_i} = {\left[ 
{{{\tilde \upsilon }_i}},{{{\tilde \omega }_i}} \right]^T}$. By substituting the designed controller in \eqref{38} and \eqref{39}, the time derivative of ${V_{6}}$ is calculated as
\begin{equation}
\begin{aligned}
       {{\dot V}_6} =& \sum\limits_{i=1}^n {\tilde z_i^T\left( {{{\dot z}_{ir}} - {{\dot z}_i}} \right)}  \\
        = & \sum\limits_{i=1}^n{\tilde z_i^T\left( {{{\dot z}_{ir}} - {\tau _i}{{\hat c}_i}  + {k_{5i}}{S_i} + {{\dot S}_{i}}} \right)}\\
        =& -\sum\limits_{i=1}^n{  \tilde z_i^T({c_{\tau i}}{{\text{sgn}(\tilde z}_i)}  - {k_{5i}}{S_i} - {{\dot S}_{i}}})\\
         \le&  - \sum\limits_{i=1}^n{\left( {{\lambda _{vi}} - {\Delta _i}} \right){\left\| {{{\tilde z}_i}} \right\|_1}},\label{44}
\end{aligned}
\end{equation}
where ${\lambda _{vi}}$ is the minimum eigenvalue of ${c_{\tau i}}$ with ${c_{\tau i}} = \text{diag}\left( 
{{c_{ai}}}, {{c_{bi}}}
 \right)$, therefore, as long as the control parameter ${\lambda _{vi}}>{\Delta_{i}}$. The proposed learning based robust controller is asymptotically stable.

\subsubsection{Step 2}
Since we have proved the asymptotic stability of the dynamic closed loop system. Then, we need to analyze the stability of the proposed kinematic controller. First, we proposed the following Lyapunov candidate function as
\begin{equation}
   {V_{7}} = \sum\limits_{i=1}^n{(\frac{1}{2}{{\tilde x}_i}^2 + \frac{1}{2}{{\tilde y}_i}^2 + \frac{1}{{{k_{2i}}}}\left( {1 - \cos {{\tilde \theta }_i}} \right) + \frac{{{k_{3i}}}}{{2{B_{1i}}}}V_{s1i}^2)}.\label{34}
\end{equation}
By substituting \eqref{29} into the time derivative of \eqref{34}, we can obtain that
\begin{equation}
    \begin{aligned}
        {{\dot V}_{7}} =  &  \sum\limits_{i=1}^n{(- {{\tilde x}_i}\upsilon _i^{cmd} + {{\tilde x}_i}{{\tilde \upsilon }_i} + {{\tilde x}_i}{\upsilon _{ir}}\cos {{\tilde \theta }_i} + {{\tilde y}_i}{\upsilon _{ir}}\sin {{\tilde \theta }_i} }+\\ & \frac{{\sin {{\tilde \theta }_i}}}{{{k_{2i}}}}{\omega _{ir}}-\frac{{\sin {{\tilde \theta }_i}}}{{{k_{2i}}}}\omega _i^{cmd} + \frac{{\sin {{\tilde \theta }_i}}}{{{k_{2i}}}}{{\tilde \omega }_i}  - \frac{{{k_{3i}}{\Lambda _{1i}}}}{{{B_{1i}}}}V_{s1i}^2 +\\& {{\tilde x}_i}{V_{s1i}}   + {{\tilde x}_i}{\Omega _{ix}} + {{\tilde y}_i}{\Omega _{iy}} + \frac{{\sin {{\tilde \theta }_i}}}{{{k_{2i}}}}{\Omega _{i\theta }}).\label{35}
    \end{aligned}
\end{equation}
We further take the controller defined in \eqref{31} into \eqref{35}, and it is calculated that 
\begin{equation}
    {{\dot V}_{7}} = {W_{1}} + {W_{2}},
\end{equation}
where ${W_1}$ and ${W_2}$ are defined as
\begin{equation}
    \begin{aligned}
        {W_{1}} =&-\sum\limits_{i=1}^n{   (\frac{{{k_{1i}}{\Lambda _{1i}}}}{{{B_{1i}}}}V_{s1i}^2 +\frac{{{k_{3i}}}}{{{k_{2i}}}}{\upsilon _{ir}}{\sin ^2}{{\tilde \theta }_i})},\\{W_{2}} =& \sum\limits_{i=1}^n{({{\tilde x}_i}{{\tilde \upsilon }_i} + {{\tilde x}_i}{\Omega _{ix}} + {{\tilde y}_i}{\Omega _{iy}} + \frac{{\sin {{\tilde \theta }_i}}}{{{k_{2i}}}}{{\tilde \omega }_i} + \sin {{\tilde \theta }_i}{\Omega _{i\theta }})}.
    \end{aligned}\label{36}
\end{equation}
\noindent\textbf{Remark 5}
It is noted that the (virtual) leader's linear velocity is set to be positive, and based on the estimation process of $\upsilon_{ir}$ defined in \eqref{13}, it is a first-order system. Therefore, by setting the initial condition start from 0, such that $\upsilon_{ir}(0)=0$, then $\upsilon_{ir}(t)\to\upsilon_r$ without overshoot as time$\to\infty$, and $\upsilon_{ir}(t)$ is ensured to be nonnegative.

Based on the definition of the bioinspired neural dynamics, we have $\Lambda_{1i}>0$ and $B_{1i}>0$; and $k_{1i}$, $k_{2i}$, and $k_{3i}$ are positive constants. Thus, we have $W_{1i}\le0$. As for the $V_{4i}$ we have the following inequalities hold
\begin{equation}
    {V_{7}} \ge \frac{1}{2}\sum\limits_{i=1}^n{\tilde x_i^2}\quad \text{and}
    \quad{V_{7}} \ge \frac{1}{2}\sum\limits_{i=1}^n{\tilde y_i^2}.
\end{equation}
We further define ${V_{7i}}$ as
\begin{equation}
     {V_{7i}} = \frac{1}{2}{{\tilde x}_i}^2 + \frac{1}{2}{{\tilde y}_i}^2 + \frac{1}{{{k_{2i}}}}\left( {1 - \cos {{\tilde \theta }_i}} \right) + \frac{{{k_{3i}}}}{{2{B_{1i}}}}V_{s1i}^2,
\end{equation}
then, the following inequalities hold as
\begin{equation}
    2{k_{2i}}{V_{7i}} \ge \left( {1 + \cos {{\tilde \theta }_i}} \right)\left( {1 - \cos {{\tilde \theta }_i}} \right) = {\sin ^2}{\tilde \theta _i}.
\end{equation}
Consequently, using Cauchy-Schwarz inequalities, the following inequalities hold as
 \begin{equation}
     \sqrt {2n{V_{7}}}  \ge \sum\limits_{i=1}^n{\left| {{{\tilde x}_i}} \right|},\quad \sqrt {2n{V_{7}}} \ge \sum\limits_{i=1}^n{\left| {{{\tilde y}_i}} \right|},\quad\text{and}, \nonumber
 \end{equation}
  \begin{equation}
\sqrt {2nk_{2m}{V_{7}}}\ge \sum\limits_{i=1}^n{\sqrt {\left( {1 + \cos {{\tilde \theta }_i}} \right){V_{7i}}}}  \ge \sum\limits_{i=1}^n{\left| {\sin {{\tilde \theta }_i}} \right|},
 \end{equation}
 where $k_{2m}$ is the maximum value of $k_{2i}$. Finally, we can make the conclusion that
 \begin{equation}
 \begin{aligned}
     W_{2}&\le({\left| {{\Omega _{ix}}} \right|_{\max }} + {\left| {{\Omega _{iy}}} \right|_{\max }} + {\left| {{{\tilde \upsilon }_i}} \right|_{\max }})\sqrt {2n{V_{7}}} \\
      &\quad + \left( {{{{\left| {{{\tilde \omega }_i}} \right|}_{\max }}} + {{\left| {{\Omega _{i\theta }}} \right|}_{\max }}} \right)\sqrt {2nk_{2m}{V_{7}}}. \label{57}
\end{aligned}
\end{equation}
 It is noticed that based on the stability analysis in \eqref{24} and \eqref{44}, we have ${{\tilde \upsilon }_i}$ and ${{\tilde \omega }_i}$, $\Omega_{ix}$, $\Omega_{iy}$, and $\Omega_{i\theta}$ are all converge to zero. Therefore, $\dot V_{4i}$ satisfies the conditions in Lemma 2. Then, based on Barbarlets lemma, we have $V_{s1i}\to0$ and $\sin{\tilde\theta_i}\to0$ as time$\to\infty$. Based on the properties of the bioinspired neural dynamics, as $V_{s1i}\to0$, then $\tilde x_i\to0$ as time$\to\infty$ as well. By considering \eqref{29}, we have $\tilde y_i\to0$ as well. Therefore, the overall system is asymptotically stable.

\section{simulation results}
\begin{figure}[h]
\centerline{\includegraphics[width=0.3\textwidth]{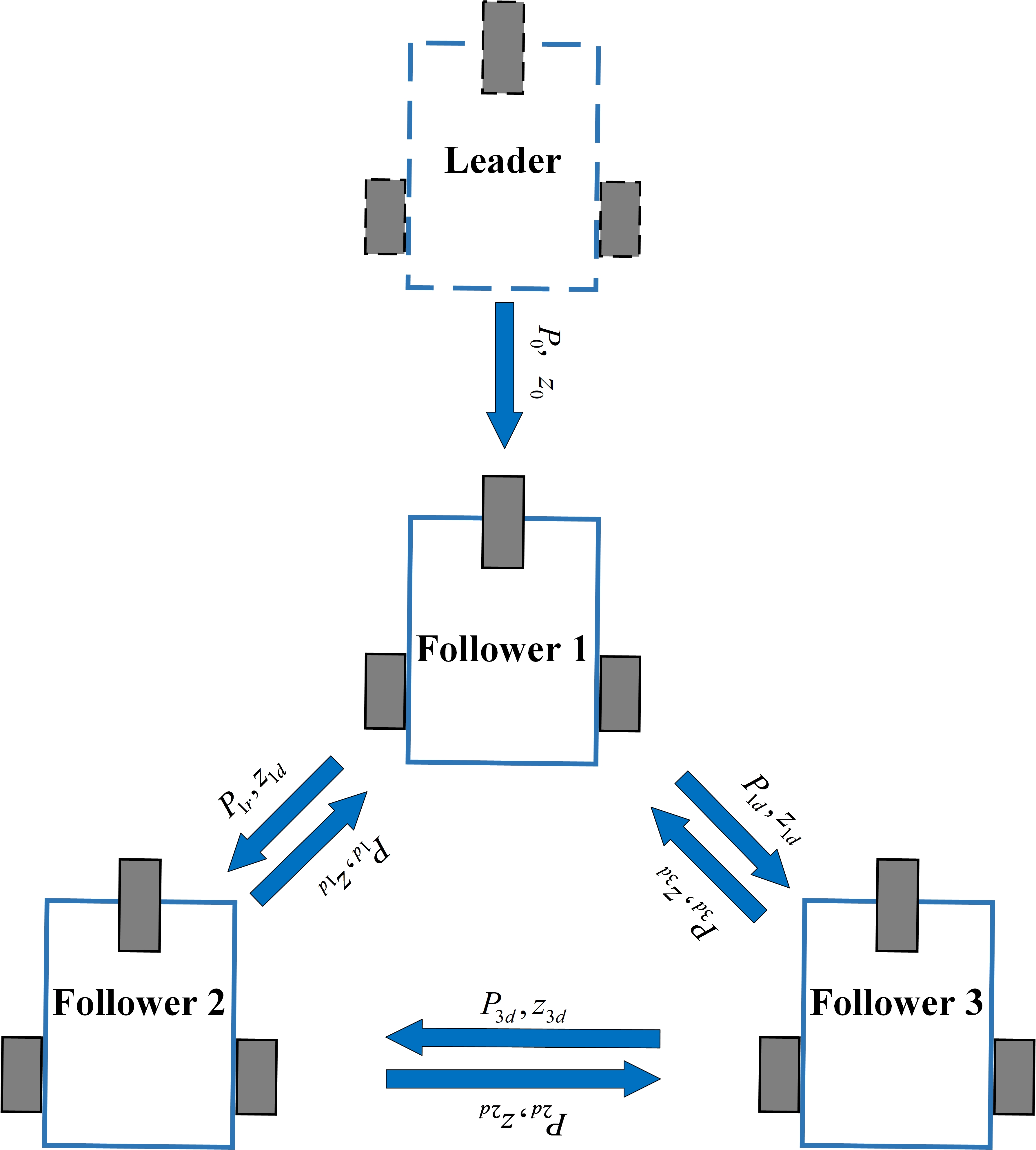}}
\caption{Communication topology of mobile robots\label{fig1}}
\end{figure}

There are multiple simulation studies have been conducted in this section to demonstrate the advantages of the designed distributed control method. First, the communication topology is demonstrated in Figure \ref{fig1}, and the relative position of each follower to the leader is given as $\Delta x_1 = 3$, $\Delta x_2 = 4$, $\Delta x_3 = 4$, $\Delta y_1 = 0$, $\Delta y_2 = 5$, $\Delta y_3 = -5$. Parameters in the distributed estimators for each robot are treated as identical, which is adjusted to ${k_{pi}} = {\left[ {15},{15},{15} \right]^T}$, $k_{a1i}=25$, and $k_{b1i} = 1$. As for the bioinspired kinematic controller, the control parameter is set to be $k_{1i}= 2$ and $k_{2i}=3$, and $k_{3i}=4$, while the bioinspired neural dynamics, $A_{i}=2$ and $B_{i}=2$. As for the learning based robust controller, $c_{ai}$ and $c_{bi}$ are set to be 3. As for the learning process, we have provided that ${{\rm{k}}_{4i}}{\rm{ =  diag}}\left( 6, 50 \right)$ and ${{\rm{k}}_{5i}}{\rm{ =  diag}}\left( 25, 50 \right)$, $d_{1i}=0.1$, and $d_{2i}=0.1\text{cos}(t)$. Mobile robot parameters are given as $a_i=0.4$ and $b_i=10$. Given the virtual leader's state as $x_r=t$ and $y_r=3+0.4\cos(-\pi/2+t)$, the respected desired velocity of the virtual leader is calculated by
\begin{equation}
    {\upsilon _r} = {\left( {{{\dot x}_r}^2 + {{\dot y}_r}^2} \right)^{0.5}}\quad \text{and}\quad {\omega _r} = \frac{{{{\ddot y}_r}{{\dot x}_r} - {{\ddot x}_r}{{\dot y}_r}}}{{{{\dot x}_r}^2 + {{\dot y}_r}^2}}.
\end{equation}
\begin{figure}[h]
\centerline{\includegraphics[width=0.4\textwidth]{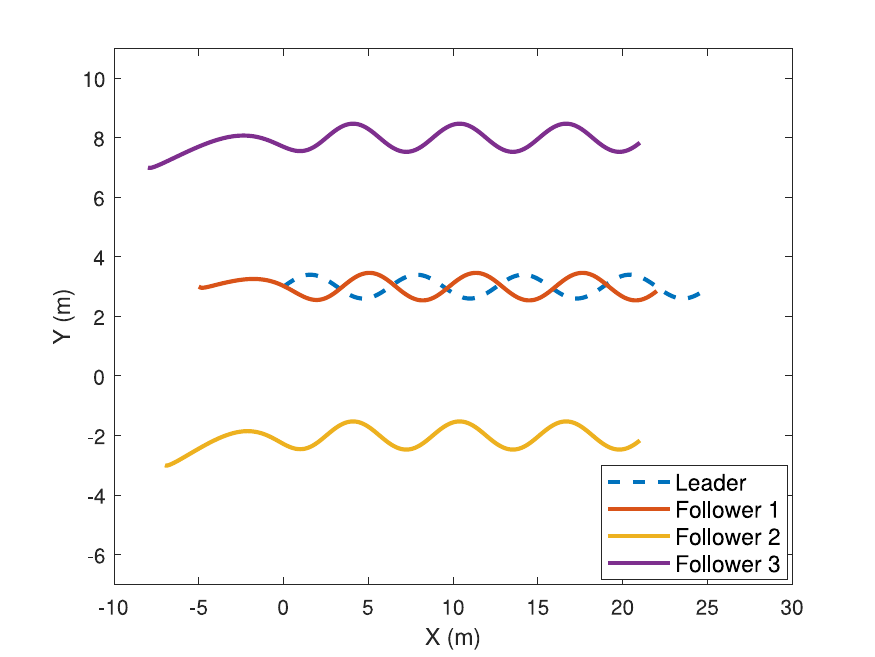}}
\caption{Trajectories of each mobile robot\label{fig2}}
\end{figure} 
The proposed method is capable of driving multiple robots to reach their relative posture within the formation, which can be seen in Figure \ref{fig2}. The bioinspired kinematic controller is capable of tracking the desired path accurately, it is seen that under the disturbances the tracking error is still capable of maintaining under small value, which can be seen in Figure \ref{fig2a}. Furthermore, the proposed distributed estimator has successfully provided the leader's state estimates, the results are provided in Figure \ref{fig3}.

\begin{figure}[h]
\centerline{\includegraphics[width=0.4\textwidth]{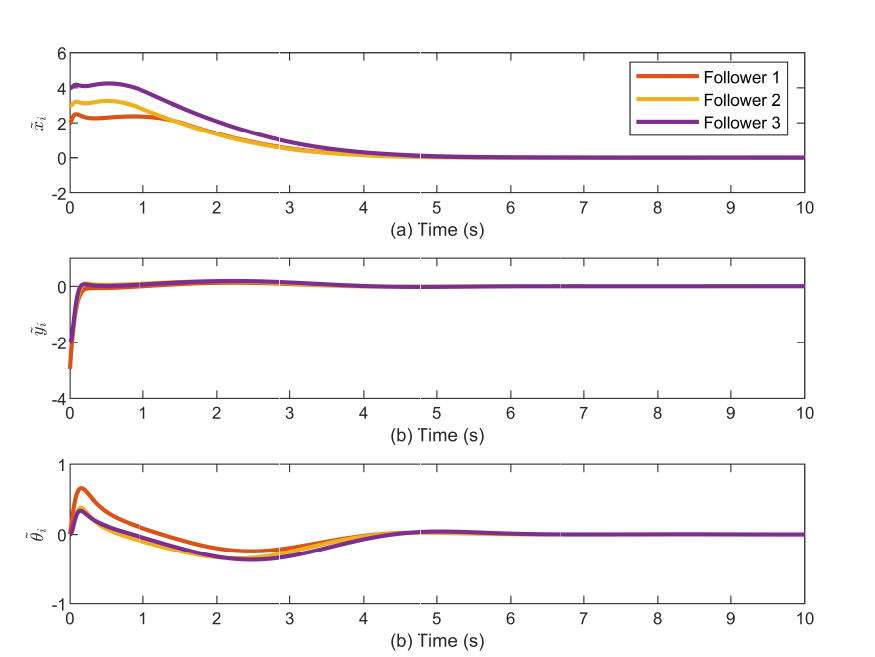}}
\caption{Tracking error of each follower\label{fig2a}}
\end{figure} 
\begin{figure}[h]
\centerline{\includegraphics[width=0.4\textwidth]{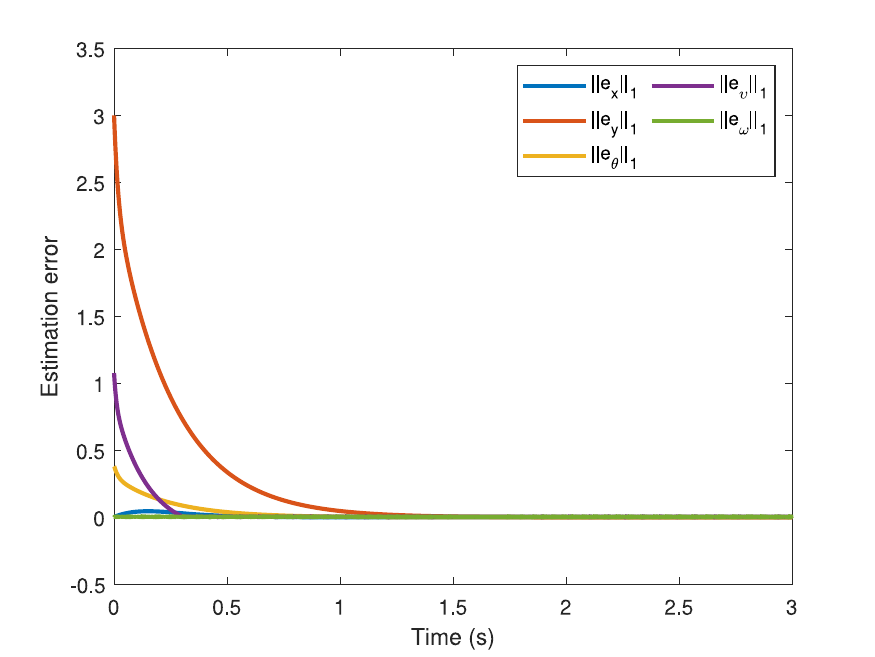}}
\caption{Estimation error of distributed estimators.\label{fig3}}
\end{figure} 
\begin{figure}[h]
\centerline{\includegraphics[width=0.45\textwidth]{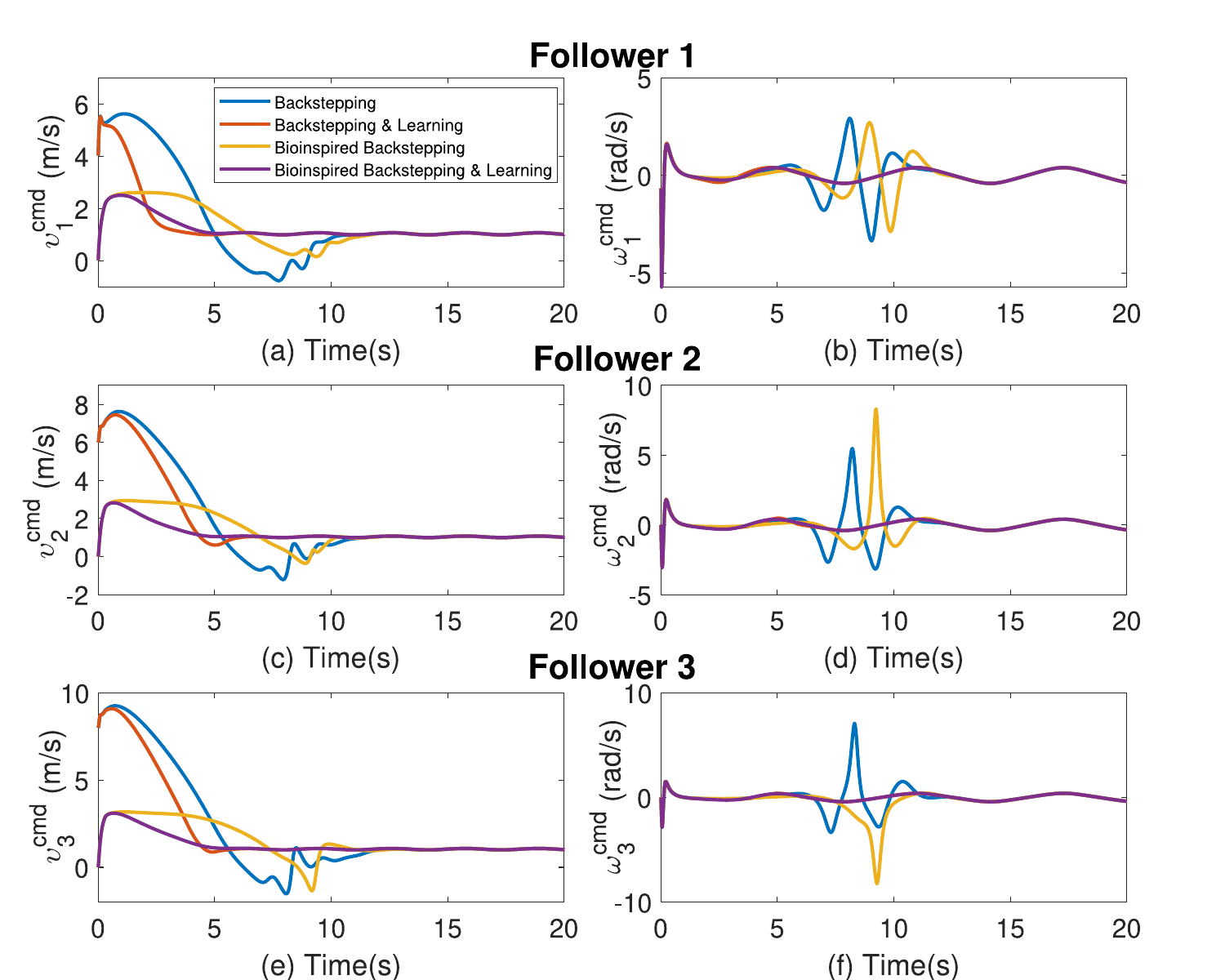}}
\caption{Velocity commands from different controllers.\label{fig4}}
\end{figure} 
The velocity commands that are generated from the kinematic controllers are seen in Figure \ref{fig4}. It is worth noting that under the disturbances and with large tracking errors, the conventional method along with the learning process is hard to tune, while the bioinspired approach provides a smoother transition of the velocity making the controller and learning process have the larger region of attraction. More importantly, the bioinspired backstepping control is capable of providing smooth velocity command without the velocity jump. As the implementation of the distributed estimator, the estimated desired velocities $\upsilon_{ir}$ and $\omega_{ir}$ both start from zero. Then, the dynamic process of the bioinspired neural dynamics makes its respected control output start from zero as well. On the other hand, the backstepping design makes the control input directly related to the tracking error, yielding an initial velocity command of $k_{1i}\tilde x_{i}$. This implies that the initial torque command will be infinitely large, making this method impractical.   Furthermore, the velocity command of the bioinspired kinematic control is strictly bounded while the maximum velocity for the backstepping control method reaches unrealistic velocity commands that are over 8 m/s for mobile robots to reach. 

\begin{figure}[h]
\centerline{\includegraphics[width=0.45\textwidth]{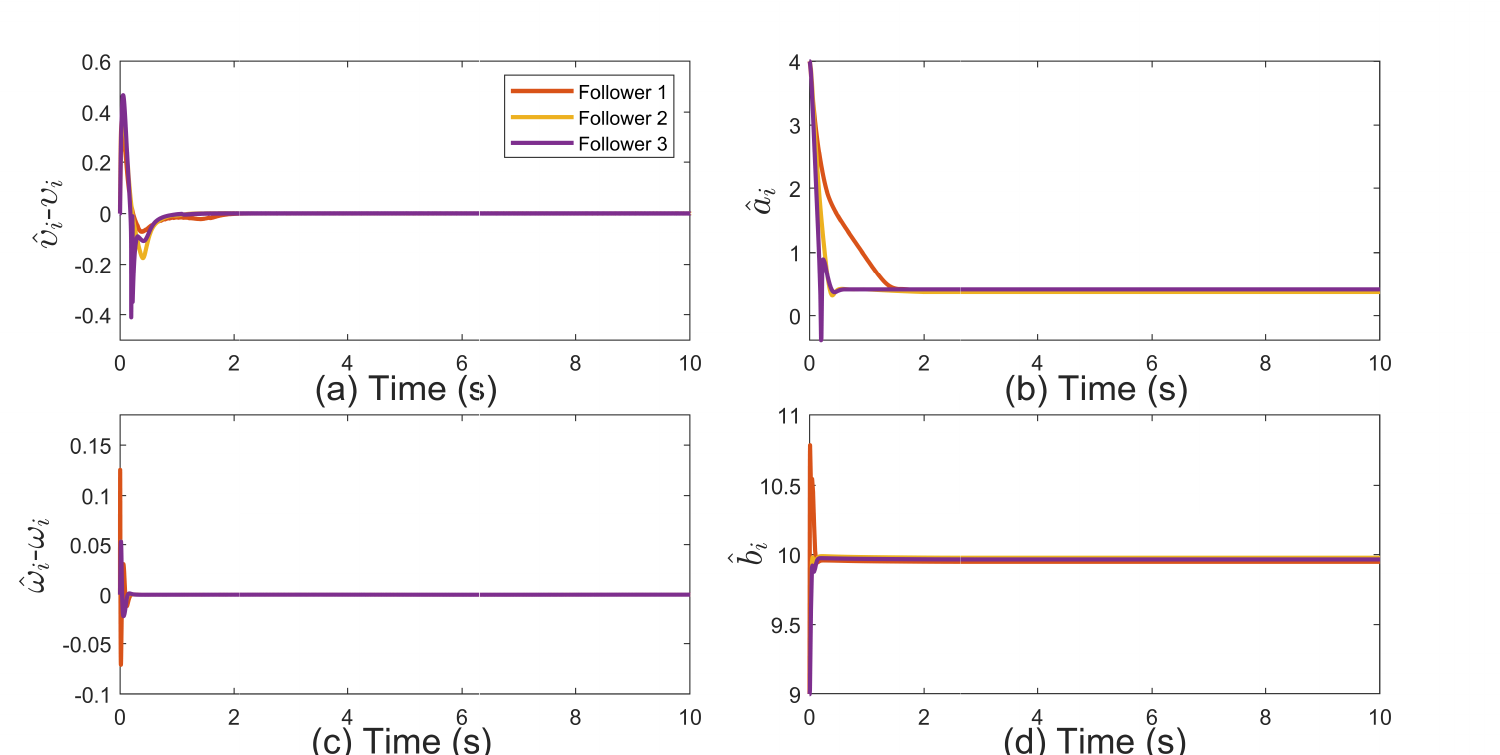}}
\caption{Learning process under disturbances.\label{fig5}}
\end{figure}

Furthermore, we have proved the effectiveness of the learning process under the disturbances. This process is critical because, in real world applications, the system may suffer from wear and tear, and its accurate model cannot be obtained. Although having a robust controller will attenuate this issue, combined with a learning process will enhance the overall performance, thus allowing a wider usage. Therefore, Figure \ref{fig5} demonstrates the convergence of the estimation error as well as the parameter estimates.
\begin{figure}[h]
\centerline{\includegraphics[width=0.45\textwidth]{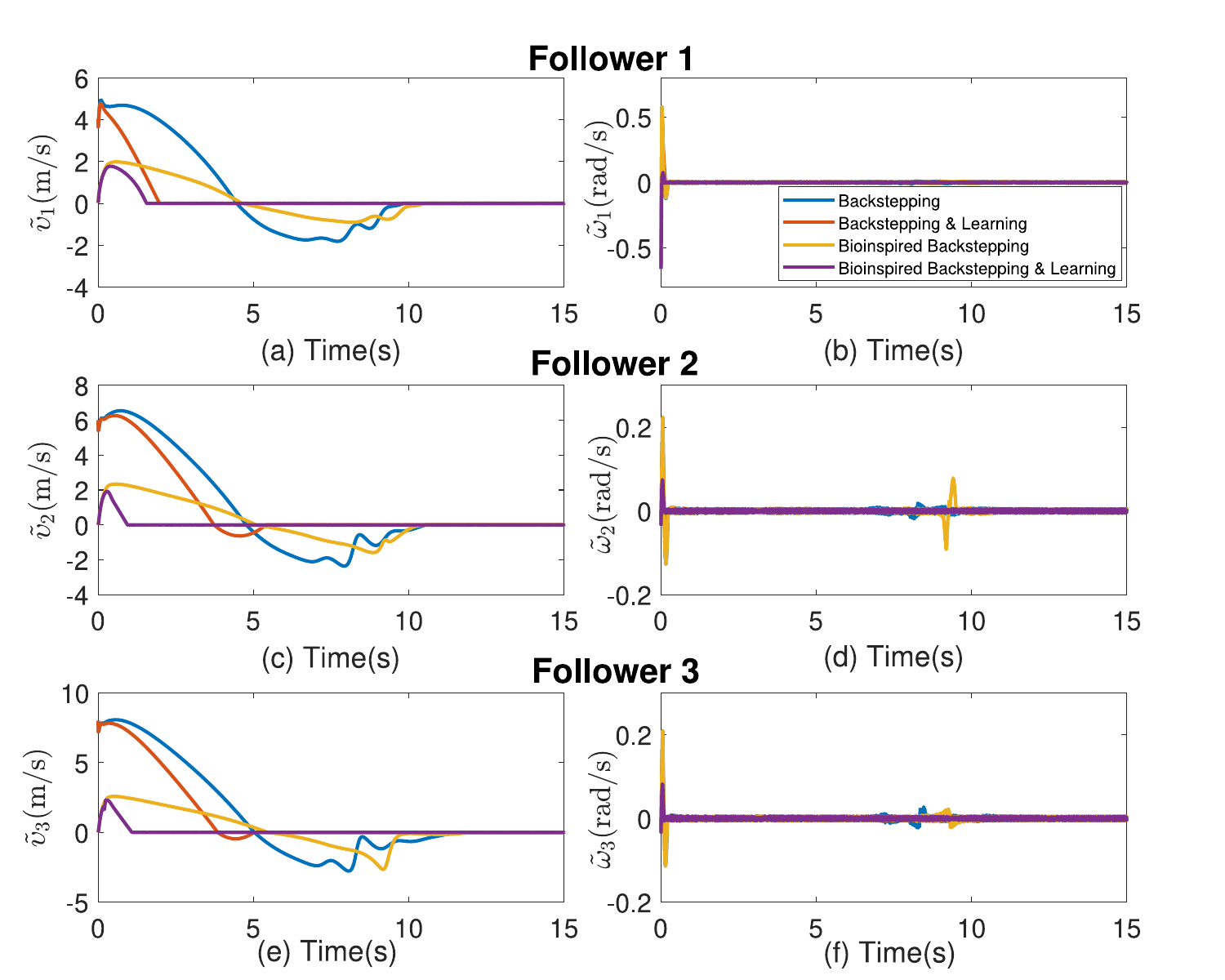}}
\caption{Velocity error under disturbances with and without learning process.\label{fig6}}
\end{figure} 
\begin{table}[!h]
\caption{Total Velocity error with different methods}
\renewcommand{\arraystretch}{1.5}
\resizebox{\columnwidth}{!}{%
\begin{tabular}{|c|c|c|c|c|}
\hline
\diagbox{Follower}{Method}     & Backstepping & \begin{tabular}[c]{@{}c@{}}Bioinspired\\  Backstepping\end{tabular} & \begin{tabular}[c]{@{}c@{}}Backstepping \& \\ Learning\end{tabular} & \begin{tabular}[c]{@{}c@{}}Bioinspired \&\\  Learning\end{tabular} \\ \hline
Follower 1 & 20.6         & 5.3                                                                & 8.9                                                                 & 1.9                                                                \\ \hline
Follower 2 & 27.2         & 15.7                                                                & 11.1                                                                & 1.1                                                                \\ \hline
Follower 3 & 33.5         & 18.9                                                                & 12.8                                                                & 1.5                                                                \\ \hline
\end{tabular}%
}
\label{tab1}
\end{table}

As for the developed adaptive bioinspired robust controller, it can be seen in Figure \ref{fig6}. Although the conventional sliding mode based approach is robust to uncertainties and disturbances, the methods without the learning process take longer to converge, with the implementation of the learning process, the control approach is less conservative, thus providing a faster convergence speed. In addition, Table \ref{tab1} further demonstrates the proposed method provides a smaller total tracking error, which also implies total energy consumption of the proposed method is smaller. Overall, multiple simulation studies have shown that the proposed learning based control design leads to better control performance.

\section{conclusion}
In this paper, the formation control problem of mobile robots under the distributed context has been addressed. The formation protocol is achieved by proposing a distributed estimator using a cascade and variable structure design technique. Then, to resolve the velocity constraints and jump issue, a distributed backstepping method is developed by taking the advantage of bioinspired neural dynamics. After that, a robust learning based dynamic control is proposed to improve the robustness and control performance. After that, the overall stability of the formation protocol is rigorously proved using the Lyapunov stability theory. Finally, various simulation studies have illustrated the performance of the proposed formation control strategy.

In the future, a deeper analysis and development of the controller design could be studied under the input to state stability analysis. A real world experimental studies could be conducted of the proposed method to verify the effectiveness of the proposed method.

\bibliographystyle{IEEEtran}
\bibliography{IEEEabrv,T-IV-24-02-0791-References}

\begin{thebibliography}{10}
\providecommand{\url}[1]{#1}
\csname url@samestyle\endcsname
\providecommand{\newblock}{\relax}
\providecommand{\bibinfo}[2]{#2}
\providecommand{\BIBentrySTDinterwordspacing}{\spaceskip=0pt\relax}
\providecommand{\BIBentryALTinterwordstretchfactor}{4}
\providecommand{\BIBentryALTinterwordspacing}{\spaceskip=\fontdimen2\font plus
\BIBentryALTinterwordstretchfactor\fontdimen3\font minus \fontdimen4\font\relax}
\providecommand{\BIBforeignlanguage}[2]{{%
\expandafter\ifx\csname l@#1\endcsname\relax
\typeout{** WARNING: IEEEtran.bst: No hyphenation pattern has been}%
\typeout{** loaded for the language `#1'. Using the pattern for}%
\typeout{** the default language instead.}%
\else
\language=\csname l@#1\endcsname
\fi
#2}}
\providecommand{\BIBdecl}{\relax}
\BIBdecl

\bibitem{Cao2023SafeRobots}
H.~Cao, H.~Xiong, W.~Zeng, H.~Jiang, Z.~Cai, L.~Hu, L.~Zhang, and W.~Lu, ``{Safe Reinforcement Learning-Based Motion Planning for Functional Mobile Robots Suffering Uncontrollable Mobile Robots},'' \emph{IEEE Transactions on Intelligent Transportation Systems}, vol.~PP, pp. 1--18, 2023.

\bibitem{Kim2023Energy-TimeRobots}
Y.~Kim and T.~Singh, ``{Energy-Time Optimal Trajectory Tracking Control of Wheeled Mobile Robots},'' \emph{IEEE/ASME Transactions on Mechatronics}, vol.~PP, pp. 1--12, 2023.

\bibitem{Sun2016TwoRobots}
W.~Sun, S.~Tang, H.~Gao, and J.~Zhao, ``{Two Time-Scale Tracking Control of Nonholonomic Wheeled Mobile Robots},'' \emph{IEEE Transactions on Control Systems Technology}, vol.~24, no.~6, pp. 2059--2069, 2016.

\bibitem{Zhou2018AgileStructures}
D.~Zhou, Z.~Wang, and M.~Schwager, ``{Agile Coordination and Assistive Collision Avoidance for Quadrotor Swarms Using Virtual Structures},'' \emph{IEEE Transactions on Robotics}, vol.~34, no.~4, pp. 916--923, 8 2018.

\bibitem{Yang2023ResearchStructure}
C.~Yang and C.~Liang, ``{Research on Multi-UAV Formation and Semi-Physical Simulation With Virtual Structure},'' \emph{IEEE Access}, vol.~11, no. November, pp. 126\,027--126\,039, 2023.

\bibitem{Wen2023Behavior-BasedComputing}
J.~Wen, J.~Yang, Y.~Li, J.~He, Z.~Li, and H.~Song, ``{Behavior-Based Formation Control Digital Twin for Multi-AUG in Edge Computing},'' \emph{IEEE Transactions on Network Science and Engineering}, vol.~10, no.~5, pp. 2791--2801, 2023.

\bibitem{Feng2023DistributedCommunication}
Y.~Feng, J.~Dong, J.~Wang, and H.~Zhu, ``{Distributed Flocking Algorithm for Multi-UAV System Based on Behavior Method and Topological Communication},'' \emph{Journal of Bionic Engineering}, vol.~20, no.~2, pp. 782--796, 3 2023.

\bibitem{Hu2023PotentialAvoidance}
\BIBentryALTinterwordspacing
C.~Hu, Y.~Hua, Q.~Wang, X.~Dong, J.~Yu, and Z.~Ren, ``{Potential field-based formation tracking control for multi-UGV system with detection behavior and collision avoidance},'' \emph{Journal of the Franklin Institute}, vol. 360, no.~17, pp. 13\,284--13\,317, 2023. [Online]. Available: \url{https://doi.org/10.1016/j.jfranklin.2023.09.060}
\BIBentrySTDinterwordspacing

\bibitem{Souza2022ModifiedEnvironments}
R.~M. J.~A. Souza, G.~V. Lima, A.~S. de~Morais, L.~C. Oliveira-Lopes, D.~C. Ramos, and F.~L. Tofoli, ``{Modified Artificial Potential Field for the Path Planning of Aircraft Swarms in Three-Dimensional Environments},'' \emph{Sensors}, vol.~22, no.~4, 2022.

\bibitem{Xu2023DistributedFilter}
Z.~Xu, T.~Yan, S.~X. Yang, and S.~A. Gadsden, ``{Distributed Leader Follower Formation Control of Mobile Robots based on Bioinspired Neural Dynamics and Adaptive Sliding Innovation Filter},'' \emph{IEEE Transactions on Industrial Informatics}, p. DOI: 10.1109/TII.2023.3272666, 2023.

\bibitem{Dai2020AdaptivePerformance}
S.~L. Dai, S.~He, X.~Chen, and X.~Jin, ``{Adaptive Leader-Follower Formation Control of Nonholonomic Mobile Robots with Prescribed Transient and Steady-State Performance},'' \emph{IEEE Transactions on Industrial Informatics}, vol.~16, no.~6, pp. 3662--3671, 2020.

\bibitem{Lu2020Neuro-AdaptiveRobots}
P.~Lu, S.~Baldi, G.~Chen, and W.~Yu, ``{Neuro-Adaptive Cooperative Tracking Rendezvous of Nonholonomic Mobile Robots},'' \emph{IEEE Transactions on Circuits and Systems II: Express Briefs}, vol.~67, no.~12, pp. 3167--3171, 2020.

\bibitem{Peng2013Leader-followerApproach}
\BIBentryALTinterwordspacing
Z.~Peng, G.~Wen, A.~Rahmani, and Y.~Yu, ``{Leader-follower formation control of nonholonomic mobile robots based on a bioinspired neurodynamic based approach},'' \emph{Robotics and Autonomous Systems}, vol.~61, no.~9, pp. 988--996, 2013. [Online]. Available: \url{http://dx.doi.org/10.1016/j.robot.2013.05.004}
\BIBentrySTDinterwordspacing

\bibitem{Miao2022Low-ComplexityFeedback}
Z.~Miao, H.~Zhong, Y.~Wang, H.~Zhang, H.~Tan, and R.~Fierro, ``{Low-Complexity Leader-Following Formation Control of Mobile Robots Using Only FOV-Constrained Visual Feedback},'' \emph{IEEE Transactions on Industrial Informatics}, vol.~18, no.~7, pp. 4665--4673, 2022.

\bibitem{Yan2022ConsensusControl}
T.~Yan, Z.~Xu, and S.~X. Yang, ``{Consensus Formation Control for Multiple AUV Systems Using Distributed Bioinspired Sliding Mode Control},'' \emph{IEEE Transactions on Intelligent Vehicles}, vol.~8, no.~2, pp. 1081--1092, 2022.

\bibitem{Zhang2023ParetoRobots}
C.~Zhang, L.~Zhou, and Y.~Li, ``{Pareto Optimal Reconfiguration Planning and Distributed Parallel Motion Control of Mobile Modular Robots},'' \emph{IEEE Transactions on Industrial Electronics}, vol.~PP, pp. 1--10, 2023.

\bibitem{Shao2022DistributedSampling}
X.~Shao, J.~Zhang, and W.~Zhang, ``{Distributed Cooperative Surrounding Control for Mobile Robots With Uncertainties and Aperiodic Sampling},'' \emph{IEEE Transactions on Intelligent Transportation Systems}, vol.~23, no.~10, pp. 18\,951--18\,961, 2022.

\bibitem{Yao2023GuidingRobots}
W.~Yao, H.~G. De~Marina, Z.~Sun, and M.~Cao, ``{Guiding Vector Fields for the Distributed Motion Coordination of Mobile Robots},'' \emph{IEEE Transactions on Robotics}, vol.~39, no.~2, pp. 1119--1135, 2023.

\bibitem{Liu2023DistributedApproach}
P.~Liu, Y.~Hao, Q.~Wang, and G.~Chen, ``{Distributed formation control of networked mobile robots from the Udwadia–Kalaba approach},'' \emph{International Journal of Robust and Nonlinear Control}, vol.~33, no.~18, pp. 11\,518--11\,537, 2023.

\bibitem{Lu2020FormationEstimators}
P.~Lu, H.~Wang, F.~Zhang, W.~Yu, and G.~Chen, ``{Formation Control of Nonholonomic Mobile Robots Using Distributed Estimators},'' \emph{IEEE Transactions on Circuits and Systems II: Express Briefs}, vol.~67, no.~12, pp. 3162--3166, 2020.

\bibitem{Moorthy2022DistributedApproach}
\BIBentryALTinterwordspacing
S.~Moorthy and Y.~H. Joo, ``{Distributed leader-following formation control for multiple nonholonomic mobile robots via bioinspired neurodynamic approach},'' \emph{Neurocomputing}, vol. 492, pp. 308--321, 2022. [Online]. Available: \url{https://doi.org/10.1016/j.neucom.2022.04.001}
\BIBentrySTDinterwordspacing

\bibitem{Liu2023FormationTechnique}
W.~Liu, X.~Wang, and S.~Li, ``{Formation Control for Leader-Follower Wheeled Mobile Robots Based on Embedded Control Technique},'' \emph{IEEE Transactions on Control Systems Technology}, vol.~31, no.~1, pp. 265--280, 2023.

\bibitem{Lin2021AdaptiveConstraints}
J.~Lin, Z.~Miao, H.~Zhong, W.~Peng, Y.~Wang, and R.~Fierro, ``{Adaptive Image-Based Leader-Follower Formation Control of Mobile Robots with Visibility Constraints},'' \emph{IEEE Transactions on Industrial Electronics}, vol.~68, no.~7, pp. 6010--6019, 2021.

\bibitem{Liu2023SecureAttacks}
Z.~Q. Liu, X.~Ge, H.~Xie, Q.~L. Han, J.~Zheng, and Y.~L. Wang, ``{Secure Leader{\&}{\#}x2013;Follower Formation Control of Networked Mobile Robots Under Replay Attacks},'' \emph{IEEE Transactions on Industrial Informatics}, vol.~PP, pp. 1--11, 2023.

\bibitem{Peng2020MobileVehicles}
X.~Peng, Z.~Sun, K.~Guo, and Z.~Geng, ``{Mobile Formation Coordination and Tracking Control for Multiple Nonholonomic Vehicles},'' \emph{IEEE/ASME Transactions on Mechatronics}, vol.~25, no.~3, pp. 1231--1242, 2020.

\bibitem{Li2020RobustFormation}
Z.~Li, Y.~Yuan, F.~Ke, W.~He, and C.~Y. Su, ``{Robust Vision-Based Tube Model Predictive Control of Multiple Mobile Robots for Leader-Follower Formation},'' \emph{IEEE Transactions on Industrial Electronics}, vol.~67, no.~4, pp. 3096--3106, 2020.

\bibitem{Wang2021DistributedAttacks}
W.~Wang, Z.~Han, K.~Liu, and J.~L{\"{u}}, ``{Distributed adaptive resilient formation control of uncertain nonholonomic mobile robots under deception attacks},'' \emph{IEEE Transactions on Circuits and Systems I: Regular Papers}, vol.~68, no.~9, pp. 3822--3835, 2021.

\bibitem{Lashkari2020DevelopmentCapability}
N.~Lashkari, M.~Biglarbegian, and S.~X. Yang, ``{Development of a novel robust control method for formation of heterogeneous multiple mobile robots with autonomous docking capability},'' \emph{IEEE Transactions on Automation Science and Engineering}, vol.~17, no.~4, pp. 1759--1776, 2020.

\bibitem{Miao2018DistributedRobots}
Z.~Miao, Y.~H. Liu, Y.~Wang, G.~Yi, and R.~Fierro, ``{Distributed Estimation and Control for Leader-Following Formations of Nonholonomic Mobile Robots},'' \emph{IEEE Transactions on Automation Science and Engineering}, vol.~15, no.~4, pp. 1946--1954, 2018.

\bibitem{Li2020ReinforcementRobots}
S.~Li, L.~Ding, H.~Gao, Y.~J. Liu, N.~Li, and Z.~Deng, ``{Reinforcement Learning Neural Network-Based Adaptive Control for State and Input Time-Delayed Wheeled Mobile Robots},'' \emph{IEEE Transactions on Systems, Man, and Cybernetics: Systems}, vol.~50, no.~11, pp. 4171--4182, 2020.

\bibitem{Hodgkin1952ANerve}
A.~Hodgkin and {Huxley}, ``{A quantitative description of membrane current and its application to conduction and excitation in nerve},'' \emph{The Journal of physiology}, vol. 117, no.~4, pp. 500--544, 1952.

\end{thebibliography}
\begin{IEEEbiography}[{\includegraphics[width=1in,height=1.25in,clip,keepaspectratio]{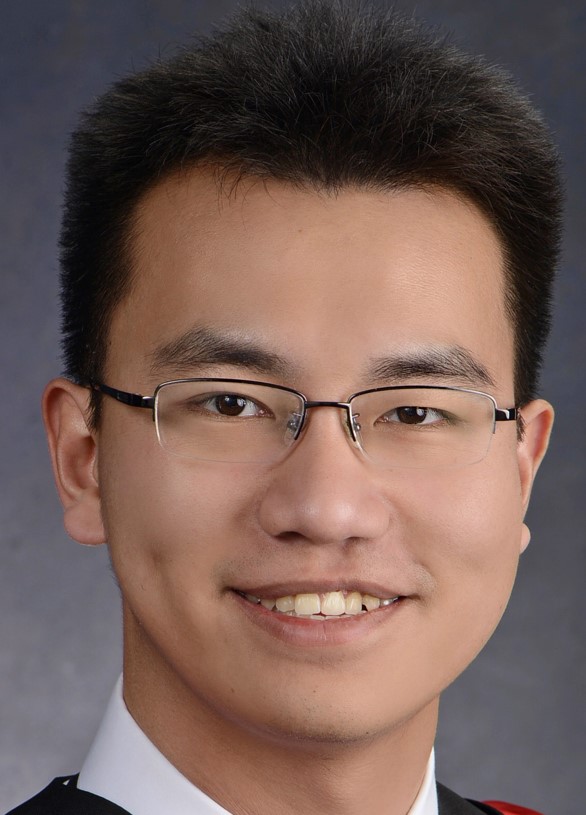}}]{Zhe Xu} (Member, IEEE) received B.ENG. degree in Mechanical Engineering in 2018, and M.A.Sc. and Ph.D. degree in Engineering Systems and Computing in 2019 and 2023, respectively, from the University of Guelph. He is currently a post-doctoral fellow with the Department of Mechanical Engineering at McMaster University. His research interests include networked systems, tracking control, estimation theory, robotics, and intelligent systems.
\end{IEEEbiography}
\begin{IEEEbiography}[{\includegraphics[width=1in,height=1.25in,clip,keepaspectratio]{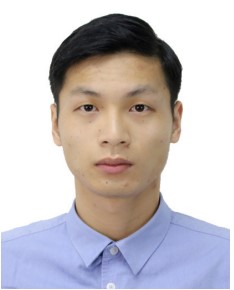}}]{Tao Yan} (Graduate Student Member, IEEE) received the B.Sc. degree from
the North China Institute of Aerospace Engineering,
Langfang, China, in 2016; the M.Sc. degree from
the Zhejiang University of Technology, Hangzhou,
China, in 2020. He is currently pursuing his Ph.D.
degree at the University of Guelph, ON, Canada.
His research interests include intelligent control,
distributed control and optimization, and networked
underwater vessel systems.
\end{IEEEbiography}
\begin{IEEEbiography}[{\includegraphics[width=1in,height=1.25in,clip,keepaspectratio]{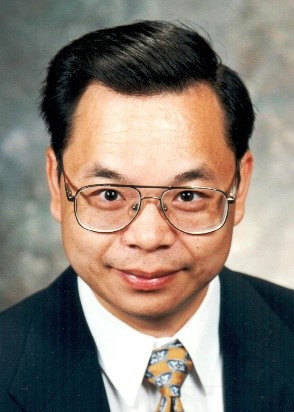}}]{Simon X. Yang} (Senior Member, IEEE) received the B.Sc. degree in engineering physics from Beijing University, Beijing, China, in 1987, the first of two M.Sc. degrees in biophysics from the Chinese Academy of Sciences, Beijing, in 1990, the second M.Sc. degree in electrical engineering from the University of Houston, Houston, TX, in 1996, and the Ph.D. degree in electrical and computer engineering from the University of Alberta, Edmonton, AB, Canada, in 1999.  Dr. Yang is currently a Professor and the Head of the Advanced Robotics and Intelligent Systems Laboratory at the University of Guelph. His research interests include robotics, intelligent systems, sensors and multi-sensor fusion, wireless sensor networks, control systems, machine learning, fuzzy systems, and computational neuroscience. 

Prof. Yang has been very active in professional activities. He serves as the Editor-in-Chief of International Journal of Robotics and Automation, and an Associate Editor of IEEE Transactions on Cybernetics, IEEE Transactions of Artificial Intelligence. 
\end{IEEEbiography}
\begin{IEEEbiography}[{\includegraphics[width=1in,height=1.25in, clip,keepaspectratio]{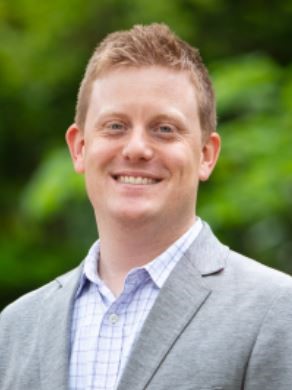}}]{S. Andrew Gadsden} (Senior Member, IEEE) earned his Bachelors in Mechanical Engineering and Management at McMaster University. He then completed his PhD in Mechanical Engineering at McMaster in the area of estimation theory and mechatronics. Andrew is an Associate Professor in Mechanical Engineering at McMaster University with a research focus in control and estimation theory, and applied artificial intelligence and machine learning. Before joining McMaster University, he was an Associate/Assistant Professor at the University of Guelph and an Assistant Professor in the Department of Mechanical Engineering at the University of Maryland, Baltimore County (USA). Andrew works/worked with a number of colleagues in NASA, the US Army Research Laboratory, US Department of Agriculture, National Institute of Standards and Technology, and the Maryland Department of the Environment. He is an elected Fellow of ASME, is a Senior Member of IEEE, and is a Professional Engineer of Ontario. Andrew is also a reviewer for a number of ASME and IEEE journals and international conferences.

\end{IEEEbiography}
\begin{IEEEbiography}[{\includegraphics[width=1in,height=1.25in, clip,keepaspectratio]{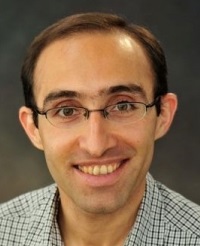}}]{Mohammad Biglarbegian} (Senior Member, IEEE) received his BSc degree (Hons.) from the University of Tehran in Tehran, Iran. He earned his MASc from the University of Toronto (Ontario, Canada) and his PhD from the University of Waterloo (Ontario, Canada). He is currently an Associate Professor with the
Department of Mechanical and Aerospace Engineering at Carleton University (Ottawa, Ontario) and is the Director of the Autonomous and Intelligent Control for Vehicles (AICV) Research Laboratory. Prior to Carleton, he was an Associate and Assistant Professor in the School of Engineering at the University of Guelph. His research interests include advanced control, robotics, estimation, and optimization for different mechatronics systems; particularly advanced autonomous robotics and vehicles. Mohammad received the University of Guelph Research Excellence Award in 2018. He currently serves as the associate editors for two international journals of Robotics and Automation as well
as Intelligence and Robotics.
\end{IEEEbiography}
\end{document}